%% file: main.tex
\documentclass[journal]{IEEEtran}
\usepackage[hidelinks, bookmarks=false, pdfauthor={Yuelong Li},
    pdftitle={},
    pdfsubject={},
    pdfkeywords={},
    pdfproducer={},
    pdfcreator={}]{hyperref}
\usepackage{cite}
\usepackage[export]{adjustbox}
\usepackage{graphicx} 

\usepackage{amsfonts}
\usepackage{bm}
\usepackage[cmex10]{amsmath}
\usepackage{algorithm, algorithmic}
\usepackage[caption=false,font=footnotesize]{subfig}
\usepackage{color}
\usepackage{verbatim}

\def\bC{\ensuremath{{\bf C}}}
\def\bD{\ensuremath{{\bf D}}}
\def\bX{\ensuremath{{\bf X}}}
\def\bE{\ensuremath{{\bf E}}}
\def\bF{\ensuremath{{\bf F}}}
\def\bA{\ensuremath{{\bf A}}}
\def\bB{\ensuremath{{\bf B}}}
\def\bW{\ensuremath{{\bf W}}}
\def\bR{\ensuremath{{\bf R}}}

\def\bM{\ensuremath{{\bf M}}}
\def\cO{\ensuremath{\mathcal{O}}}
\def\bx{\ensuremath{{\bf x}}}
\def\bp{\ensuremath{{\bf p}}}
\def\by{\ensuremath{{\bf y}}}
\def\be{\ensuremath{{\bf e}}}

\def\bT{\ensuremath{{\bf T}}}

\def\bone{\ensuremath{{\bf 1}}}

\def\bJ{\ensuremath{{\bf J}}}
\def\bT{\ensuremath{{\bf T}}}
\def\bH{\ensuremath{{\bf H}}}
\def\bK{\ensuremath{{\bf K}}}
\def\bI{\ensuremath{{\bf I}}}

\def\cO{\ensuremath{{\mathcal O}}}

\def\bT{\ensuremath{{\bf T}}}
\def\bI{\ensuremath{{\bf I}}}
\def\bD{\ensuremath{{\bf D}}}
\def\bX{\ensuremath{{\bf X}}}
\def\bY{\ensuremath{{\bf Y}}}
\def\bu{\ensuremath{{\bf u}}}
\def\bv{\ensuremath{{\bf v}}}
\def\bx{\ensuremath{{\bf x}}}

\def\bE{\ensuremath{{\bf E}}}
\def\bA{\ensuremath{{\bf A}}}
\def\bS{\ensuremath{{\bf S}}}

\def\vec{\ensuremath{{\mathrm{vec}}}}

\def\cP{\ensuremath{{\mathcal P}}}

\def\bLambda{\ensuremath{\bm{\Lambda}}}
\def\bbeta{\ensuremath{\bm{\beta}}}

\begin{document}

\title{A Maximum A Posteriori Estimation Framework for Robust High Dynamic Range Video Synthesis}

\author{Yuelong~Li,~\IEEEmembership{Student~Member,~IEEE,}
    Chul~Lee,~\IEEEmembership{Member,~IEEE,}
    and~Vishal~Monga,~\IEEEmembership{Senior~Member,~IEEE}
    \thanks{Y. Li and V. Monga are with the Department
        of Electrical Engineering, Pennsylvania State University, University Park,
    PA, 16802 USA (e-mails: yul200@psu.edu and vmonga@engr.psu.edu).}
    \thanks{C. Lee is with the Department of Computer Engineering, Pukyong National University, Busan 48513, Korea (e-mail: chullee@pknu.ac.kr).}
}

\maketitle

\begin{abstract}
    High dynamic range (HDR) image synthesis from multiple low dynamic range (LDR) exposures continues to be actively researched. The extension to HDR video synthesis is a topic of significant current interest due to potential cost benefits. For HDR video, a stiff practical challenge presents itself in the form of accurate correspondence estimation of objects between video frames. In particular, loss of data resulting from poor exposures and varying intensity make conventional optical flow methods highly inaccurate. We avoid exact correspondence estimation by proposing a statistical approach via maximum a posterior (MAP) estimation, and under appropriate statistical assumptions and choice of priors and models, we reduce it to an optimization problem of solving for the foreground and background of the target frame. We obtain the background through rank minimization and estimate the foreground via a novel multiscale adaptive kernel regression technique, which implicitly captures local structure and temporal motion by solving an unconstrained optimization problem. Extensive experimental results on both real and synthetic datasets demonstrate that our algorithm is more capable of delivering  high-quality HDR videos than current state-of-the-art methods, under both subjective and objective assessments. Furthermore, a thorough complexity analysis reveals that our algorithm achieves better complexity-performance trade-off than conventional methods.
\end{abstract}

\begin{IEEEkeywords}
    High dynamic range video, maximum a posterior estimation, kernel regression.
\end{IEEEkeywords}

\IEEEpeerreviewmaketitle

\input{section1}

\input{section2}

\input{section3}

\input{section4}

\input{section5}

\input{section6}

\appendices

\input{appendix1}

\input{appendix2}


\bibliographystyle{IEEEtran}
\bibliography{hdr_image,hdr_video,super}








\end{document}

%% file: section1.tex
\section{Introduction}
\IEEEPARstart{H}{igh} dynamic range (HDR) images, with several orders higher
dynamic range than conventional low dynamic range (LDR) images, are capable of
faithfully carrying full information of natural
scenes~\cite{HDRI_BOOK_Reinhard}.  This advantage has brought HDR imaging into
a wide range of application areas, including game and movie
industries~\cite{HDRI_BOOK_Reinhard}. However, digital imaging devices designed
for directly capturing HDR images are generally prohibitively expensive to be
employed in practical
applications~\cite{nayar_adaptive_2003,tocci_2011_TOG,kronander2014unified}.
An alternative approach is to merge multiple images of the same scene taken
with varying exposure times~\cite{debevec_recovering_2008}. However, directly
merging multiple LDR frames is prone to bringing in ghosting artifacts.
Therefore, to address such an issue, a lot of ghost-free HDR imaging techniques
have been
proposed~\cite{sen_robust_2012,hu_hdr_2013,lee_ghost-free_2014,oh_robust_2014}.
For a comprehensive survey on this literature, we refer the readers
to~\cite{CGF_Tursun_2015,HDRI_BOOK_Reinhard}.

\begin{figure}
    \centering
    \subfloat[] {\includegraphics[width=4.2cm]{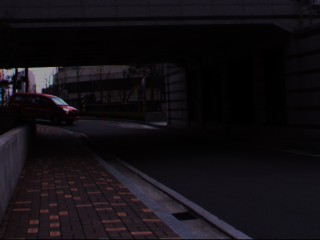}}\,\!
    \subfloat[] {\includegraphics[width=4.2cm]{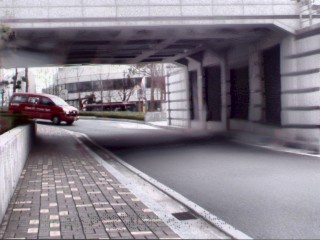}}\\
    \caption{
        A simple extension of the state-of-the-art HDR imaging algorithm~\cite{hu_hdr_2013} to HDR video synthesis fails to produce high-quality HDR frames. (a) The original LDR frame, (b) HDR frame synthesized by Hu {\it et al}.'s HDR imaging algorithm~\cite{hu_hdr_2013}. Video result available at \url{https://youtu.be/K_MkSxMURXo}.
    }
    \label{fig:hu}
\end{figure}

Similar to HDR image synthesis, HDR video synthesis has also become an
active research topic, as there is a growing demand for HDR videos.
The transition from HDR imaging to HDR video seems straightforward.
Instead of capturing multiple images of the same scene with varying exposures,
one can alternate the exposure times between successive frames, and then
synthesize several consecutive frames together repeatedly on a frame basis to
form an HDR video~\cite{kang_high_2003}. However, despite the popularity of HDR
imaging, only a limited number of algorithms have been proposed to synthesize
HDR videos through the aforementioned approach.  Although we may apply the
existing ghost-free HDR imaging techniques to HDR video synthesis through the
sliding window approach, such a simple extension fails to provide high-quality
HDR video frames. Fig.~\ref{fig:hu} shows such an example, in which the
state-of-the-art ghost-free HDR imaging algorithm~\cite{hu_hdr_2013} suffers from visually
objectionable artifacts. The major challenge in HDR video
synthesis is to establish accurate correspondences between successive frames
with different exposures. Specifically, the violation of the constant intensity
assumption made typically by optical flow methods \cite{lucas_ijcai_1981,liu2009beyond} together with the presence of under-exposed or saturated regions
imposes great challenges on the computation of optical flow.  Moreover, since
the input LDR video frames are typically taken with two to three different exposure
times, unlike the case in HDR image synthesis where multiple exposure
times are allowed, content loss due to under-exposure or saturation is even
more severe. As a consequence, direct application of ghost-free HDR image
synthesis techniques to HDR video usually yields poor results.

Due to the aforementioned challenges, a vast majority of research on HDR video
focuses on hardware-based
approaches~\cite{tocci_2011_TOG,kronander2014unified,unger2007high}. However,
these methods need specially designed devices and are typically prohibitively
expensive. Therefore, software-based approaches via synthesis of alternatively
exposed LDR frames still have their merits, as they only require conventional
cameras. Recent work has focused on addressing HDR video synthesis challenges at
an algorithmic level.  In~\cite{kang_high_2003}, Kang {\it et al}.
compensated the exposure differences between frames by adjusting their
exposures through the camera response function, and then applied conventional
optical flow techniques to estimate correspondences. However, their method
relies heavily on the accuracy of the optical flow estimation, and it usually
fails when complex motions exist.  To increase the robustness of optical flow,
Mangiat and Gibson~\cite{mangiat_high_2010} developed a block-based motion
estimation technique instead of the conventional gradient-based
ones~\cite{lucas_ijcai_1981,horn_mit_1980}. They further alleviated
registration errors by post-processing with a modified bilateral filter, called
HDR filter~\cite{mangiat_spatially_2011}, and applied it to large scale
security and surveillance~\cite{mangiat_milcom_2011}.
In~\cite{kalantari_patch-based_2013}, Kalantari {\it et al}. composed optical
flow based registration with patch-based synthesis~\cite{sen_robust_2012} to
handle complex motions. In their method, temporal coherency is enhanced by
optical flow, while incorrect correspondences are corrected by patch-based
synthesis. Although their algorithm usually provides high-quality HDR video
sequences, they may distort moving objects and introduce color artifacts.

{\bf Motivations and Contributions:} The stiff challenge in HDR video synthesis is in the exact correspondence estimation between input  frames taken with different exposures.
To address the challenge, we avoid the correspondence estimation by formulating HDR video synthesis
statistically as a MAP estimation problem, which subsequently reduces to an
optimization problem that implicitly captures motion information. Specifically,
each video frame is separated into a static background and a sparse foreground,
which can be synthesized via rank minimization and multiscale adaptive kernel
regression, respectively. Compared with conventional approaches, our algorithm
\emph{for the first time} introduces MAP estimation approach, as well as kernel
regression method into the literature of HDR video synthesis. We call it
MAP-HDR. Further, we improve kernel regression to make it locally adaptable in handling motion and dealing with missing (saturated/dark) samples by solving an unconstrained optimization problem.
Our major contributions in this work are
summarized as follows.\footnote{Preliminary results of this work have been
presented in part in~\cite{ICIP_Li_2015}. In this paper, we provide more
thorough theoretical derivation of the MAP estimation problem and develop a
new kernel regression method for accurate correspondence estimation.
Furthermore, more comprehensive experiments are included, including objective quality assessment and complexity analysis.}


\begin{enumerate}
    \item \textbf{A novel MAP estimation framework}: We develop, for the first
    time, a MAP estimation framework for HDR image and video
    synthesis.  It can be shown that many existing techniques
    in~\cite{lee_ghost-free_2014,oh_robust_2014} can be regarded as
    specific instances of this framework.  Furthermore, based on reasonable
    statistical assumptions and appropriate choice of statistical models, we
    can reduce the general MAP estimation framework towards a specific
    optimization problem, whose solution provides synthesized HDR video frames.

    \item \textbf{Multiscale adaptive kernel regression approach}: We develop a
    novel multiscale adaptive kernel regression approach. In this work, we
    apply it to the synthesis of foreground radiance data, as part of our
    MAP-HDR algorithm.  Its effectiveness in accurate HDR data estimation is
    demonstrated through extensive experiments.

    \item \textbf{Experimental validation and insight}: We evaluate the merits of
    the MAP-HDR algorithm on both real-world and synthetic datasets.
    Experimental results demonstrate that our algorithm is capable of producing
    high-quality artifact-free HDR videos both under subjective and objective
    assessments, particularly in the face of complex motion when many state-of-the-art
    methods fail.

    \item \textbf{Complexity analysis}: We carry out a thorough complexity
    analysis over the MAP-HDR algorithm as well as the state-of-the-art HDR
    video synthesis methods. Our algorithm turns out to be of comparable
    complexity with state-of-the-art methods, while providing higher-quality
    results. Thus, our algorithm achieves a better complexity-performance
    trade-off.

    \item \textbf{Reproducibility}: All the datasets and the MATLAB code for
    producing the experimental results in this paper are publicly available
    online on our project website.\footnote{\url{http://signal.ee.psu.edu/hdrvideo.html}}
\end{enumerate}

The rest of the paper is organized as follows: Section~\ref{sec:skr} provides a succinct review of various kernel regression techniques. Section~\ref{sec:derivation} describes the general formulation of HDR video synthesis through MAP estimation and develops a novel multiscale adaptive kernel regression approach to estimate foreground data. Section~\ref{sec:implementation} discusses implementation
details of the proposed MAP-HDR algorithm. Section~\ref{sec:experiments} shows the effectiveness of
the MAP-HDR algorithm by comparison against several state-of-the-art methods both subjectively and
objectively, and a computational complexity analysis is provided. Finally, Section~\ref{sec:conclusion} concludes this paper.

%% file: section2.tex
\section{Preliminaries of Kernel Regression}
\label{sec:skr}

For the sake of completeness, we briefly review the classical kernel regression (CKR) and its variants in~\cite{takeda_kernel_2007, takeda_super-resolution_2009}, on which the proposed algorithm is based.
CKR refers to the technique of minimum mean squared error (MMSE) estimation of
regression function through local polynomial approximation. Suppose that we aim
to estimate the regression relationship $z(\cdot)$ between a set of data
samples $\{y_i\in\mathbb{R}\}_{i=1}^P$ and their locations
$\{\bx_i\in\mathbb{R}^d\}_{i=1}^P$, given by
\begin{equation}
    y_i = z(\bx_i) + \varepsilon_i, \quad i = 1,\dots, P.
\end{equation}

Assuming that $z(\cdot)$ is smooth, by the Taylor's formula, we can
approximately expand it locally as a polynomial around some point $\bx$ near
the sample location $\bx_i$, {\it i.e.},
\begin{equation}
    z(\bx_i)\approx z(\bx) + \nabla z^T(\bx_i - \bx) + \frac{1}{2}(\bx_i - \bx)^T\mathcal{H}(z)(\bx_i - \bx) + \cdots,
\end{equation}
where $\nabla z$ and $\mathcal{H}(z)$ denote the gradient and Hessian of
$z(\cdot)$ at $\bx$, respectively. By parametrizing its polynomial expansion,
the estimation of $z(\cdot)$ can be simplified as the following weighted linear
least square problem:
\begin{equation}
    \min_{\{\bm{\beta}_k\}}\;\sum_{i=1}^P[y_i - \beta_0 - \bm{\beta}_1^T(\bx_i - \bx) - \cdots]^2K_h(\bx_i - \bx),
    \label{eqn:wls}
\end{equation}
where $K_h(\cdot)$ is the \emph{kernel function} for imposing weights on
each squared error term. Its parameter $h$ is called \emph{bandwidth}, which
controls the extent of smoothing. A commonly used kernel function is the
Gaussian kernel, given by
\begin{equation}
    K_h(\bu) = \frac{1}{(2\pi)^{\frac{d}{2}}h^d}\exp\!\left(-\frac{\|\bu\|^2}{2h^2}\right)\!.
\end{equation}

In~\cite{takeda_kernel_2007}, Takeda {\it et al.} developed an iterative
steering kernel regression (ISKR) scheme in 2D and showed that it can be
successfully applied to various image processing applications, including 
denoising, interpolation, and super-resolution. Further,
in~\cite{takeda_super-resolution_2009}, it was shown that state-of-the-art
performance can be achieved in video super-resolution by generalizing ISKR to
3D, called 3D ISKR. The major advantage of ISKR over CKR is that the kernel
function $K(\cdot)$ of ISKR is data-adaptive, {\it i.e.}, it has the following
form:
\begin{equation}
    K_{\bH_i}(\bu) = \frac{\exp\!\left(-\frac{1}{2}\bu^T\bH_i^{-1}\bu\right)}{(2\pi)^{\frac{d}{2}}\sqrt{\det(\bH_i)}},
    \label{eqn:skr}
\end{equation}
where $\bH_i$ is called the \emph{steering matrix}. At each sample location,
the steering matrix is estimated from the structure tensor
matrix~\cite{schaal1998constructive} with some modifications on its singular
values to avoid singularity and flatten its directionality, which would
otherwise be too strong. By substituting $K_h(\cdot)$ in (\ref{eqn:wls}) with
$K_{\bH_i}(\cdot)$ and solving the weighted least square problem, we can obtain
the optimal coefficients $\widehat{\bm{\beta}}$, given by
\begin{equation}
    \widehat{\bm{\beta}} = (\bX^T\bK\bX)^{-1}\bX^T\bK\by,
    \label{eqn:coeff}
\end{equation}
where $\by = [y_1, y_2, \dots, y_P]^T$ is the data sample vector, and $\bm{\bK}
= \mathrm{diag}\{K_{\bH_i}(\bx_i - \bx)\}_{i=1}^P$ is the diagonal weighting
matrix with $K_{\bH_i}(\bx_i-\bx)$ on the diagonal. For video processing, the
data samples are collected from a local spatial-temporal 3D block, and
typically the center of the block is the target pixel to be estimated. Also,
$\bX$, called \emph{design matrix}, is given by
\begin{equation}
    \bX =
    \begin{bmatrix}
        1 & (\bx_1 - \bx)^T & \mathrm{vech}^T\{(\bx_1 - \bx)(\bx_1 - \bx)^T\} & \dots\\
        1 & (\bx_2 - \bx)^T & \mathrm{vech}^T\{(\bx_2 - \bx)(\bx_2 - \bx)^T\} & \dots\\
        \vdots & \vdots & \vdots & \vdots \\
        1 & (\bx_P - \bx)^T & \mathrm{vech}^T\{(\bx_P - \bx)(\bx_P - \bx)^T\} & \dots
    \end{bmatrix},
    \label{eqn:design}
\end{equation}
where $\rm vech(\cdot)$ denotes the half-vectorization operator of the
lower-triangular portion of a symmetric matrix~\cite{takeda_kernel_2007}.

%% file: section3.tex
\section{Proposed Algorithm (MAP-HDR)}
\label{sec:derivation}
\subsection{Image Acquisition Model}
\label{ssec:physical}

\begin{figure}
    \centering
    \includegraphics[scale=0.9]{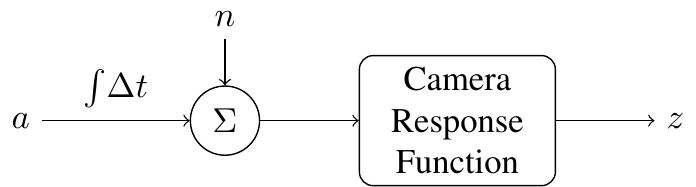}
    \caption{Physical model for HDR image acquisition.}
    \label{fig:physical}
\end{figure}

We first present an image acquisition model, on which the MAP-HDR algorithm is
based. The charge collected by a sensor equals the product of the irradiance
value $a$ and the exposure time $\Delta t$. Then, the resulting exposure value
($a\Delta t$) is corrupted by Gaussian random noise $n$. Finally, the pixel
intensity value $z$ is obtained by the camera response function
$g(\cdot)$~\cite{debevec_recovering_2008}, a nonlinear mapping from the
exposure value to the pixel intensity value.  Fig.~\ref{fig:physical} depicts
this physical process; more formally, this process can be expressed by
\begin{equation}
    z = g(a\Delta t+n).
\end{equation}
For a well-exposed pixel, the camera response function is invertible at its
radiance level. Then, the conditional probability density of $z$ given the
knowledge of $a$ is given by
\begin{align}
    f(z|a) &= \frac{d}{dz}\mathrm{Pr}\{g(a\Delta t + n) < z\}\nonumber\\
    &= \frac{d}{dz}\mathrm{Pr}\{a\Delta t+n<g^{-1}(z)\}\nonumber\\
    &= \frac{d}{dz}\Phi\!\left(\frac{g^{-1}(z) - a\Delta t}{\sigma_n}\right)\! \propto \exp\!\left(-\frac{(a - \widehat{a})^2}{2\sigma_n^2/\Delta t^2}\right)\!,
    \label{eqn:gauss}
\end{align}
where $\sigma_n$ and $\Phi$  denote the standard deviation of $n$ and the
cumulative distribution function (CDF) of the standard Gaussian distribution,
respectively, and $\widehat{a}=g^{-1}(z)/\Delta t$ is the estimated irradiance
value.

Suppose we are given $N$ consecutive frames, taken with alternating exposures,
{\it i.e.}, long and short. Each frame consists of $K$ pixels.  We separate
each frame into foreground and background regions. To this end, we first define
a binary matrix $\bS \in \{0,1\}^{K \times N}$ that supports foreground regions
as
\begin{equation}
    S_{ij} =
    \begin{cases}
        1, & \text{if pixel $(i,j)$ is in foreground},\\
        0, & \text{if pixel $(i, j)$ is in background}.
    \end{cases}
\end{equation}
Then, we develop a probabilistic model of $\bS$ based on the assumption that
the interactions of a given pixel with all the others are wholly described by
its neighbors, {\it i.e.}, it satisfies the Markov property. Then, we model
$\bS$ as a Markov random field (MRF) to exploit the spatial interactive
relationship among pixels. In this work, we employ the Ising model~\cite{blake_markov_2011}, which is
known to be simple yet effective in modeling interactions between neighboring
pixels~\cite{cevher_sparse_2010}. Specifically, the probability density
function (PDF), which can be regarded as the prior probability of $\bS$, is
given by
\begin{equation}
    f(\bS)\propto \exp \!\left(-\sum_{(i,j)\in
            {\mathcal E}}w_{ij}s_is_j - \sum_{i\in
            {\mathcal V}}\lambda_{i}s_i \right)\!,
    \label{eqn:ising}
\end{equation}
where $\mathcal V$ and $\mathcal E$ denote the sets of vertices and edges of the
graphical representation of $\bS$, respectively, and $w_{ij}$ and $\lambda_{i}$
are the edge interaction costs and vertex weights. Also, $s_i\in\{0, 1\}$
corresponds to the $i$-th pixel in $\bS$.

\subsection{MAP Estimation-based HDR Video Synthesis}
\label{ssec:formmap}

We make two assumptions on the input videos. First, scene information is included in either of alternating (exposure) frames. Second, global camera motions as well as local object motions between adjacent frames are bounded, {\it i.e.}, they can be reasonably but not arbitrarily large.

Based on these assumptions, we formulate HDR video synthesis as a MAP estimation problem based on the probabilistic model in (\ref{eqn:ising}). Let $\bD :=
[\vec(I_1), \vec(I_2), \dots, \vec(I_N)]$ denote the observation matrix of a
scene, where each $I_k$ is the observed irradiance map of the $k$-th frame
obtained by the camera response function~\cite{debevec_recovering_2008}. As
mentioned earlier, we separate each frame into foreground and
background regions to consider object motions between input frames. Let $\bF$
and $\bB$ denote matrices of the inherent foreground and background
irradiance, respectively. Then, the irradiance matrix $\bA$ for the
synthesized HDR video can be obtained by
\begin{equation}
    \bA = \cP_{\bS}(\bF) + \cP_{\bS^\mathsf{c}}(\bB),
    \label{eqn:hdr_synthesis}
\end{equation}
where $\cP_\bS (\bY)$ denotes the sampling operator defined by
$$
\left[\cP_\bS (\bY)\right]_{ij} =
\begin{cases}
    Y_{ij},  &  \mbox{if $S_{ij} = 1$}, \\
    0,       &  \mbox{otherwise}. \\
\end{cases}
$$

Note, however, that the foreground and background regions $\bF$ and $\bB$,
respectively, and the foreground support $\bS$ are unknown in
(\ref{eqn:hdr_synthesis}). Therefore, we estimate $\bB$, $\bF$, and $\bS$ given
the observation $\bD$.  Specifically, let
$f(\bS, \bB, \bF | \bD)$ denote the joint probability of $\bS$, $\bB$, and
$\bF$ given $\bD$. Then, we formulate the estimation of $\bS$, $\bB$, and $\bF$
as the following MAP estimation problem:
\begin{equation}
    (\widehat{\bS}, \widehat{\bB}, \widehat{\bF}) = \underset{\bS,\bB,\bF}{\arg\max}\; f(\bS, \bB, \bF|\bD).
    \label{eqn:map}
\end{equation}

For simplicity, we assume Lambertian surfaces, where all irradiance maps are
normalized by subtracting its minimum value and dividing by its range. Then, we can make following two independence assumptions. First, the irradiance
values depend only on the properties of the objects, and thus the irradiance values
of the foreground objects and the background scenes are statistically
independent, {\it i.e.},
\begin{equation}
    f(\cP_{\bS^\mathsf{c}}(\bY), \cP_\bS(\bY)) = f(\cP_{\bS^\mathsf{c}}(\bY)) f(\cP_\bS(\bY)),
    \label{eqn:asum1}
\end{equation}
where $\bY = \bD, \bA$. Second, we assume that the irradiance at each
pixel is independent of $\bS$, provided that we know whether it belongs to
foreground or background. In other words,
\begin{align}
    \label{eqn:asum2}
    f(\cP_\bS(\bY)|\bS) &= f(\cP_\bS(\bY)), \\
    \label{eqn:asum3}
    f(\cP_{\bS^\mathsf{c}}(\bY)|\bS) &= f(\cP_{\bS^\mathsf{c}}(\bY)).
\end{align}
Then, based on the statistical assumptions in
(\ref{eqn:asum1})--(\ref{eqn:asum3}), we can rewrite the estimation problem in
(\ref{eqn:map}) as
\begin{align}
    &(\widehat{\bS}, \widehat{\bB}, \widehat{\bF}) = \underset{\bS,\bB,\bF}{\arg\max}\; \log f(\cP_{\bS^\mathsf{c}}(\bD)|\bB) + \log f(\bB) \nonumber\\
    & + \log f(\cP_\bS(\bF)|\cP_\bS(\bD)) + \log f(\cP_{\bS}(\bD)) + \log f(\bS).
    \label{eqn:final}
\end{align}
The detailed derivation of (\ref{eqn:final}) is provided in Appendix~\ref{sec:appendix_map}.

First, recall that $\bD$ is the observation of $\bA$ with noise.  Thus,
from the observation model in~(\ref{eqn:gauss}) with the Gaussian noise, we can
approximate the first term in (\ref{eqn:final}) $\log
f(\cP_{\bS^\mathsf{c}}(\bD)|\bB)$ as negatively proportional to
$\|\cP_{\Omega}(\bD - \bB)\|_F^2$, where $\Omega$ denotes the support of
well-exposed background, given by $\Omega := \cP_{\bS^\mathsf{c}}(\bM)$, and $\|\bY\|_F:=\big(\sum_{ij}|Y_{ij}|^2\big)^{1/2}$ is the Frobenius norm of a matrix. Here,
$M_{ij} = 1$ if a pixel intensity $Z_{ij}$ satisfies $Z_{\rm
    th}<Z_{ij}<Z_{\max}-Z_{\rm th}$, where $Z_{\rm th}$ and $Z_{\max}$  are the
threshold value and maximum pixel intensity, respectively, and $M_{ij} = 0$
otherwise.  Note that we use $\Omega$ instead of $\bS^\mathsf{c}$. This is
based on the observation that the derivations in (\ref{eqn:gauss}) holds only
for well-exposed pixels and ill-exposed pixels are invalid, since they are
constant independent of $\bB$.

Second, considering that the background scene is static, we regard $f(\bB)$ as
a low-rank prior of the background scene matrix. More specifically, $f(\bB) =
\frac{1}{C}\exp(-\alpha\|\bB\|_\ast)$ with a normalization constant $C$, where
$\|\bB\|_* = \sum_k \sigma_k(\bB)$ denotes the nuclear norm, and $\sigma_k
(\bB)$ is the $k$-th largest singular value of $\bB$. Note that the low-rank
prior plays a role similar to the Laplacian prior in the structured sparsity
model~\cite{cevher_sparse_2008}.  In addition, as discussed in
Section~\ref{ssec:physical}, the last term $\log f(\bS)$ in (\ref{eqn:final})
is given by (\ref{eqn:ising}).

Third, we choose a conditional Gaussian distribution for
$\cP_\bS(\bF)|\cP_\bS(\bD)$, then $\log f(\cP_\bS(\bF)|\cP_\bS(\bD))\propto
-\|\bC^{-\frac{1}{2}}\cP_\bS(\bF-\bE)\|_F^2$, where
$\bE=\mathcal{E}[\cP_\bS(\bF)|\cP_\bS(\bD)]$, $\bC$ is the covariance
matrix of $\cP_\bS(\bF)|\cP_\bS(\bD)$. Note that $\bE$ corresponds to the
MMSE estimator of $\bF$ given $\bD$.  Also, we
choose a Laplacian prior for $f(\cP_\bS(\bD))$, {\it i.e.}, $\log
f(\cP_\bS(\bD))\propto -\|\cP_\bS(\bD)\|_1 = -\sum_i d_is_i$, where $d_i$ and
$s_i$ are the elements of $\bD$ and $\bS$, respectively. Note that, since $s_i$
is binary, an improper choice of the prior would result in an intractable
combinatorial optimization problem~\cite{TPAMI_Kolmogorov_2004}.

Then, under these simplifications, we reformulate the MAP estimation in
(\ref{eqn:final}) as the following optimization problem:
\begin{align}
    (\widehat{\bS}, \widehat{\bB}, \widehat{\bF}) &= \underset{\bS,\bB,\bF}{\arg\min}\; \frac{1}{2}\|\cP_{\Omega}(\bD-\bB)\|_F^2 + \alpha\|\bB\|_{\ast} + \beta\|\bS\|_1 \nonumber\\
    &\quad + \gamma\|\bW\vec(\bS)\|_1 + \|\bC^{-\frac{1}{2}}\vec(\cP_\bS(\bF - \bE))\|^2_F,
    \label{eqn:obj}
\end{align}
where $\bW$ is the weighting matrix that accounts for the interactions between
neighboring pixels, and $\alpha, \beta, \gamma > 0$ are constant parameters to
control the relative importance between each term.

Notice that, in~(\ref{eqn:obj}), for any fixed $\bS$, we can obtain
$\cP_\bS(\widehat{\bF}) = \cP_\bS(\widehat{\bE})$. Thus, $\bF$ can be
considered independent of other terms, and we can separate the optimization
in~(\ref{eqn:obj}) into two subproblems, {\it i.e.},

\begin{align} (\widehat{\bS},
    \widehat{\bB}) &= \underset{\bS,\bB}{\arg\min}\;
    \frac{1}{2}\|\cP_{\Omega}(\bD-\bB)\|_F^2 +
    \alpha\|\bB\|_{\ast}\nonumber\\
    &{}\qquad +
    \beta\|\bS\|_1 + \gamma\|\bW\vec(\bS)\|_1
    \label{eqn:obj_deghost}
\end{align}
and
\begin{equation}
    \cP_{\widehat{\bS}}(\widehat{\bF}) =
    \mathcal{E}\!\left[\cP_{\widehat{\bS}}(\bF)|\cP_{\widehat{\bS}}(\bD)\right]\!.
\end{equation}
A solution to the subproblem in (\ref{eqn:obj_deghost}) can be obtained by the
rank minimization for HDR deghosting~\cite{lee_ghost-free_2014}. Then, it
remains to find
$\mathcal{E}\!\left[\cP_{\widehat{\bS}}(\bF)|\cP_{\widehat{\bS}}(\bD)\right]$.
In this work, we propose to employ kernel regression~\cite{takeda_kernel_2007}, which
is a widely-used MMSE approach. In~\cite{takeda_super-resolution_2009}, kernel
regression has been proven to be effective in implicitly capturing motion
information and interpolating missing pixels, thus rendering high-quality video
results. Let us now describe how to extend the standard kernel
regression~\cite{takeda_kernel_2007} to deal with HDR data.

\subsection{Locally Adaptive Kernel Regression via Optimization}
\label{ssec:lakr}

Since HDR video synthesis can be regarded as a video reconstruction problem,
{\it i.e.}, recovery of data in saturated or under-exposed regions, kernel
regression is a natural choice to be considered. However, direct application of
kernel regression techniques, including the classical ones and their variants,
to HDR video synthesis may fail to generate high-quality videos, for the
following reasons:
\begin{enumerate}

    \item HDR data are usually noisy and grainy after exposure adjustment, which degrades the performance of kernel regression. Specifically, reliable estimation of the steering matrices in such regions is generally difficult. \label{item:challenge1}

    \item Some regions in HDR frames may be missing due to saturation or under-exposure. The polynomial approximation of kernel regression may fail to represent underlying data faithfully in such regions, especially where a significant portion of pixels are missing. \label{item:challenge2}

    \item HDR data varies significantly in intensity values, and even after exposure adjustment the intensity differences between frames may still be significant, making the data in different frames biased.\label{item:challenge3}
\end{enumerate}

\begin{figure}
    \centering
    \subfloat[Short exposure]{\includegraphics[width=0.475\columnwidth]{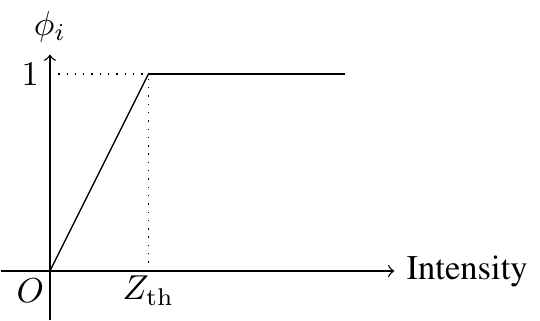}}\,
    \subfloat[Long exposure]{\includegraphics[width=0.475\columnwidth]{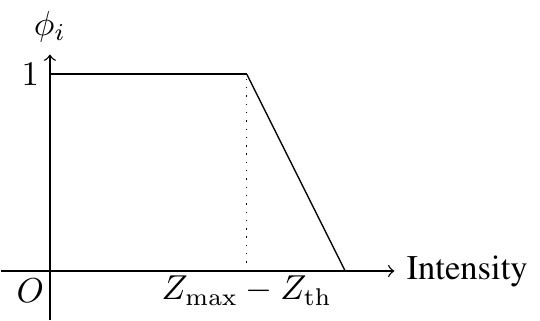}}

    \caption{Plots of weighting function $\phi_i$.}
    \label{fig:radiomteric}
\end{figure}

We develop a locally adaptive kernel regression method to address the aforementioned issues for HDR video synthesis. Firstly, to address \ref{item:challenge1}), we impose a radiometric weight $\nu_i$ for pixel $i$, defined by $\nu_i=\phi_i\omega_i$, where $\phi_i$ assigns lower weights to less properly exposed pixels as depicted in Fig.~\ref{fig:radiomteric}. For the center pixel index $c$ to be estimated, $\omega_i$ is defined as
\begin{equation}
    \omega_i=
        \begin{cases}
            e^{-\sigma(y_i-y_c)_+^2}, & \text{if reference frame is long exposure,} \\
            e^{-\sigma(y_c-y_i)_+^2}, & \text{otherwise,}
        \end{cases}
\end{equation}
with $\sigma=\kappa(1-w_c)$ for some positive constant $\kappa$ and $x_+:=\max\{x,0\}$. When the center pixel is saturated, $\omega_i$ encourages fitting towards higher values, but towards lower values when the center pixel is under-exposed.

Then, we reformulate the weighted least square problem in~(\ref{eqn:wls}) as
\begin{equation}
    \min_{\bm{\beta}} \;\; (\by - \bX\bm{\beta})^T\bm{\Lambda}(\by - \bX\bm{\beta})+\varepsilon\|\bm{\beta}\|_2^2,
    \label{eqn:swls}
\end{equation}
where $\by$ is the data sample vector, and $\bm{\Lambda} =
    \mathrm{diag}\{K_{\bH}(\bx_i - \bx)\nu_i\}_{i=1}^P$ is the weighting matrix.
Note that the \textit{Tikhonov regularization} term $\|\bm{\beta}\|_2^2$ is added in (\ref{eqn:swls}) to prevent overfitting, and $\varepsilon>0$ is the regularization parameter to control its
effect. Then, the optimal coefficients can be obtained as a closed-form solution
to~(\ref{eqn:swls}), given by
\begin{equation}
    \widehat{\bm{\beta}} = (\bX^T\bm{\Lambda}\bX + \varepsilon\bI)^{-1}\bX^T\bm{\Lambda}\by.
    \label{eqn:betaopt}
\end{equation}
In addition, we note that the computation of the structure tensor matrix is intractable in practice. Therefore, instead of explicitly computing the structure tensor
matrix and using it to obtain the steering matrix, we
determine it implicitly by minimizing the estimation error. More specifically, we try to find $\bH$ that minimizes the weighted least squares error
\begin{equation}
    (\by - \bX\widehat{\bm{\beta}})^T\bm{\Lambda}(\by -
    \bX\widehat{\bm{\beta}}) +
    \varepsilon\|\widehat{\bm{\beta}}\|_2^2.
    \label{eqn:mwls}
\end{equation}

Since the covariance
matrix must be symmetric positive definite (SPD), if we
directly solve for it, a SPD constraint has to be imposed
that is hard to deal with. Thus, we represent the inverse of steering
matrix using its \textit{Cholesky factorization} instead, given by
\begin{equation}
    \bH = \bR^\ast\bR,
\end{equation}
where $\bR\in\mathbb{R}^{3\times 3}$ is
some upper triangular matrix. In addition to removing the
SPD constraint, using this representation can also bring
down the dimensionality of the problem since $\bR$ is upper
triangular.  However, simply minimizing~(\ref{eqn:mwls})
may cause overfitting. In particular, $\bR$ may tends to
either $0$ or infinity. In the former case, the kernel
would be highly spread out, making the estimate
undesirably blurred, while in the latter case, the
obtained Gaussian kernel would be highly concentrated
around $\bx$, making the estimate strongly biased towards
$z(\bx)$.
To prevent the former case, we use
the \emph{unnormalized} Gaussian kernel
$\exp\left(-\frac{1}{2}\|\bR\bu\|_2^2\right)$.
To avoid the latter
case, \textit{i.e.}, to prevent $\bR$ tending infinity, we
propose to add the regularization term $\|\bR\|_F^2$.  In addition, since $\|\bR\|_F^2 = \mathrm{tr}(\bR^*\bR) =
\|\bH\|_\ast^2$, minimizing it jointly with~(\ref{eqn:mwls})
would effectively \emph{elongate} and \emph{rotate} the
covariance matrix $\bH$ along the local edge directions. As
mentioned in Section~\ref{sec:skr}, this is the major
characteristics of the steering matrix, the one we aim to
estimate.

Secondly, we address~\ref{item:challenge2}) by modifying the polynomial fitting model in~(\ref{eqn:wls}) as
\begin{equation}
    y_i=
    \begin{cases}
        \beta_0 + \bm{\beta}_1^T(\bx_i-\bx) + \dots, & \text{if pixel $i$ is well-exposed,}\\
        t, & \text{otherwise,}
    \end{cases}
\end{equation}
where $t=Z_{\mathrm{th}}$ when the reference frame is of lower exposure, and $t=Z_{\mathrm{max}}-Z_{\mathrm{th}}$ otherwise.
Accordingly, the following cost term is added to~(\ref{eqn:swls})
\begin{equation}
    \sum_{i=1}^P(1-\nu_i)K_\bH(\bx_i-\bx)(y_i-t)^2.
    \label{eqn:saturationcost}
\end{equation}
The cost term in~(\ref{eqn:saturationcost}) discourages $\bH$ from steering towards dark pixels when the center pixel is saturated and toward bright pixels when the center pixel is under-exposed.

Finally, to address~\ref{item:challenge3}), we apply kernel regression to boosted~\cite{kang_high_2003} and Gamma-adjusted pixel values that are of less variations than the irradiance values. We extrapolate it when applying the camera response function to avoid data truncation.

Combining all the terms, we can formulate the estimation of the steering matrix
as the following optimization problem:
\begin{align}
    \min_{\bR} \;\; &(\by - \bX\widehat{\bm{\beta}}_\bR)^T\bm{\Lambda}_\bR(\by - \bX\widehat{\bm{\beta}}_\bR) \nonumber\\
    &{} \;\; + (\by-t\bone)^T\widetilde{\bm{\Lambda}}_{\bR}(\by-t\bone)+\varepsilon\|\widehat{\bm{\beta}}_\bR\|_2^2 + \lambda\|\bR\|_F^2,
    \label{eqn:objective}
\end{align}
where $\widetilde{\bm{\Lambda}}_{\bR}:=\mathrm{diag}\{K_\bH(\bx_i-\bx)(1-\nu_i)\}$, $\varepsilon$ and $\lambda$ are positive regularization parameters, and $\bone\in\mathbb{R}^P$ is the vector containing all ones. Here, we explicitly put the
subscript $\bR$ to indicate the dependence on $\bR$, {\it i.e.}, $\bm{\Lambda}$, $\widetilde{\bm{\Lambda}}$, $\bK$, and $\widehat{\bm{\beta}}$ are functions of $\bR$.
As $\varepsilon$ or $\lambda$ gets increased, the
data becomes more smooth in general, as they play a role similar
to that of bandwidth selection in CKR.
However, the performance of our algorithm is less sensitive
to the choice of $\varepsilon$ and $\lambda$ compared with the sensitivity of bandwith
selection in CKR.


%% file: section4.tex
\section{Implementation}
\label{sec:implementation}

\subsection{Performance and Speed Improvements}
Note that the cost function in~(\ref{eqn:objective}) is smooth but nonconvex.
Therefore, to solve the optimization, we employ the
Broyden-Fletcher-Goldfarb-Shanno (BFGS)
algorithm~\cite{code_nlopt,nocedal_mc_1980,liu_mp_1989,nocedal2006numerical},
which is well-known for its fast convergence rate and robustness.
In addition, we determine the gradient of~(\ref{eqn:objective})
analytically via matrix calculus techniques, instead of approximating it
numerically through finite differences. This can greatly speedup the
optimization process, since significantly fewer matrix multiplications and
inversions are required. Specifically, let $\mathcal{C}$ denote the cost
function in~(\ref{eqn:objective}), then, in
Appendix~\ref{sec:appendix_gradient}, we derive its gradient, which is given by
\begin{align}
    \nabla\mathcal{C} &= \mathcal{UT}\Bigg\{2\bR\sum_{i,j}k_jz_j(\bx_j-\bx)(\bx_j-\bx)^T \nonumber\\
    &\qquad {} \times\left(k_iz_iJ_{ij}-\lambda\widehat{\beta}_iT_{ij}\right)+2\mu\bR\nonumber\\
    &\quad {} - \bR\!\sum_{i}\left(k_iz_i^2+\tilde{k}_i\tilde{z}_i^2\right)(\bx_i-\bx)(\bx_i-\bx)^T\Bigg\},
    \label{eqn:gradient}
\end{align}
where $\bT:=(\bX^T\bLambda\bX+\varepsilon\bI)^{-1}\bX^T$, $\bJ:=\bX\bT$, $k_i=\Lambda_{ii}$, $\tilde{k}_i=\widetilde{\Lambda}_{ii}$, $z_i:=y_i-\hat{y}_i$, $\tilde{z}_i:=y_i-t$, and
$\widehat{y}_i$ are estimates of $y_i$ obtained by
solving~(\ref{eqn:swls}). Also, $\mathcal{UT}\{\cdot\}$ denotes the operator to
extract the upper triangular part of a matrix.

In~\cite{takeda_super-resolution_2009}, second order polynomial model is
employed both in spatial and temporal axes, and five to seven consecutive
frames were shown to be helpful in preserving the details. However, in HDR
video synthesis, usually only two to three frames are available. In this
scenario, a second order polynomial has high risk of introducing
large oscillations, producing visually noticeable artifacts. Therefore, we
limit the order of polynomial to one (linear) in our implementation.

\subsection{Pyramidal Framework for Handling Large Motions}

\begin{figure*}[t]
    \centering
    \includegraphics[scale=0.9]{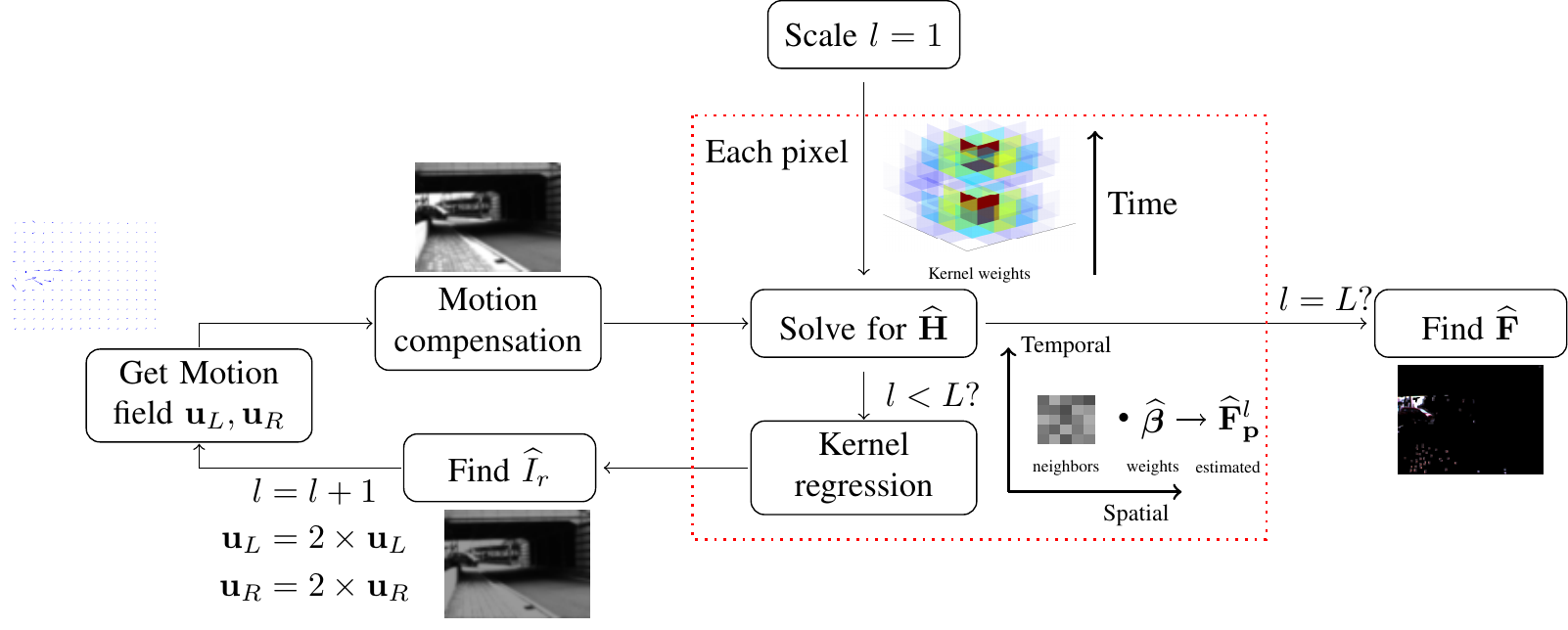}
    \caption{The flowchart of multiscale kernel regression. We demonstrate its intermediate steps using one sample taken from the~\emph{ParkingLot} dataset.~\cite{lee_rate-distortion_2012}}
    \label{fig:flowchart}
\end{figure*}

As kernel regression is performed on a small local block, it may fail
to find correct correspondences for large motion, degrading the synthesis
performance by introducing ghosting artifacts. To address such a limitation, we
propose a pyramidal implementation of the aforementioned kernel regression
method. Specifically, we first build an $L$-level Gaussian pyramid of the input frame,
indexed from coarse to fine. At the $l$-th level, we first obtain an estimation
of the reference frame. We apply the proposed locally adaptive kernel
regression algorithm to obtain the foreground part, while filling in the background part
with the scaled down version of the background matrix $\widehat{\bB}$, obtained
by solving~(\ref{eqn:obj_deghost}). Then, we compute the motion
fields among the estimated reference frame and its original neighboring frames
using optical flow~\cite{liu2009beyond,brox_high_2004,bruhn2005lucas}. After
the estimation process, exposure differences can be effectively alleviated, and
hence motion estimation becomes reliable.
We then scale up to the
$(l+1)$th level and warp the neighboring frames towards the reference one using
the previously obtained motion fields. On the finest level, we
synthesize the warped irradiance frames using the aforementioned technique.
Finally, to lower down the number of iterations in
solving~(\ref{eqn:objective}), we
initialize all $\bH$'s at the top level with identity matrices, which corresponds to
CKR, and $\bH$'s in $(l+1)$th level with those propagated from $l$th level.
This strategy usually brings the
initial $\bH$'s closer to the optimal points, providing faster convergence. The
overall flowchart of the proposed pyramidal implementation is depicted in
Fig.~\ref{fig:flowchart}

\begin{algorithm}[t]
    \renewcommand{\algorithmicrequire}{\textbf{Input:}}
    \renewcommand{\algorithmicensure}{\textbf{Output:}}
    \caption{MAP-HDR Algorithm}

    \begin{algorithmic}[1]
        \REQUIRE $I_{r-1}$, $I_r$, $I_{r+1}$, ${\bf M}$, $\alpha$, $\beta$, $\gamma$, $\kappa$, $\lambda$, $\mu$, $L$
        \STATE Initialize $\bS_1 \gets {\bf M}$, $\bD\gets[\vec(I_{r-1}),\vec(I_r),\vec(I_{r+1})]$, $\bu_L\gets\mathbf{0}$, $\bu_R\gets\mathbf{0}$, and $k=1$.
        \WHILE{stopping criteria not met}
        \STATE $\Omega\gets \cP_{\bS_k^\mathsf{c}}(\bM)$.
        \STATE $\bB_{k+1} \gets \underset{\bB}{\arg \min} \;\frac{1}{2} \|\mathcal{P}_{\Omega}(\bD - \bB) \|_F^2 + \alpha \|\bB\|_*$.
        \STATE $\bS_{k+1} \gets \underset{\bS\in\{0,1\}^{K\times N}}{\arg \min} \;\frac{1}{2} \|\mathcal{P}_{\Omega}(\bD - \bB_{k+1}) \|_F^2 + \beta \|\bS\|_1\newline
        \mbox{\hspace{1.25cm}} + \gamma \|{\bf W}{\rm vec}(\bS)\|_1$.
        \STATE $k \gets k+1$.
        \ENDWHILE
        \STATE $\widehat{\bB}\gets \bB_k$, $\widehat{\bS}\gets \bS_k$.
        \STATE $\{\widehat{\bB}^l\}_{l=1}^L\gets \mathrm{Pyr}(\widehat{\bB})$, $\{\widehat{\bS}^l\}_{l=1}^L\gets \mathrm{Pyr}(\widehat{\bS})$,\newline
        $\{\bD^l\}_{l=1}^L\gets\mathrm{Pyr}(\bD)$, $\{\bM^l\}_{l=1}^L\gets\mathrm{Pyr}(\bM)$.
        \FOR{$l=1$ \TO $L$}
        \STATE $\bu_L\gets2\times\bu_L$, $\bu_R\gets2\times\bu_R$.
        \STATE $\widehat{I}_{r-1}^l\gets\mathcal{W}_{\bu_L}(I_{r-1}^l)$, $\widehat{I}_{r+1}^l\gets\mathcal{W}_{\bu_R}(I_{r+1}^l)$.
        \FORALL{$\bp\in\mathrm{Supp}\{\bS^l\}$}
        \STATE Find $\widehat{\bH}$ by solving~(\ref{eqn:objective}).
        \STATE Find $\widehat{\bbeta}$ from~(\ref{eqn:betaopt}).
        \STATE $\widehat{\bF}^l_\bp\gets\be_1^T\widehat{\bbeta}$.
        \ENDFOR
        \STATE $\widehat{I}^l_r\gets\cP_{\bS^l}(\widehat{\bF}^l) + \cP_{{(\bS^l})^\mathsf{c}}(\widehat{\bB}^l)$.
        \IF{$l\neq L$}
        \STATE $\bv_L\gets\mathrm{ME}(\widehat{I}^l_r, \widehat{I}^l_{r-1})$, $\bv_R\gets\mathrm{ME}(\widehat{I}^l_r, \widehat{I}^l_{r+1})$.
        \STATE $\bu_L\gets\bu_L+\bv_L$, $\bu_R\gets\bu_R+\bv_R$.
        \ENDIF
        \ENDFOR
        \ENSURE $\widehat{I}_r$
    \end{algorithmic}
    \label{alg:hdrv}
\end{algorithm}

Putting all the pieces together, our HDR video synthesis algorithm for
synthesizing one HDR frame can be summarized in Algorithm~\ref{alg:hdrv}.  We
use $\mathrm{Pyr}(\bX)$ to denote the pyramidal decomposition of $\bX$ and
$\bX_{\bp}$ to denote one pixel of image $\bX$ at location $\bp$, $\bX^l$ to
denote the scaled-down version of $\bX$ in the $l$-th level, and
$\mathrm{ME}(I_1, I_2)$ to denote motion estimation between frame $I_1$ and
$I_2$.  Also, $r$ is the reference index, $I_r$, $I_{r-1}$, and $I_{r+1}$
denote the reference frame, its previous and next frames, respectively, and
$\mathcal{W}_\bu(I)$ denotes the warping operator on $I$ using motion field
$\bu$.

%% file: section5.tex
\section{Experiments and Discussions}
\label{sec:experiments}

\begin{table}
    \renewcommand{\arraystretch}{1.1}
    \centering
    \caption{Exposure times in seconds for each dataset.}

    \begin{tabular}{c||c|c|c|c}
     \hline
     \hline
             & \emph{Students} & \emph{Hallway2} & \emph{ParkingLot} & \emph{Bridge2} \\
     \hline
     \hline
     Long    & 0.005 & 0.005 &0.012 &5 \\
     \hline
     Short   & 0.0005 &0.0005 &0.001 &0.25 \\
     \hline
     \hline
    \end{tabular}
    \label{table:exposures}
\end{table}

We evaluate the performance of the proposed MAP-HDR algorithm on seven
challenging test video sequences shown in
Figs.~\ref{fig:fire}--\ref{fig:breakdown}:
\emph{Fire}~\cite{kalantari_patch-based_2013},
\emph{Bridge2}~\cite{kronander2014unified},
\emph{ParkingLot}~\cite{lee_rate-distortion_2012},
\emph{Hallway2}~\cite{kronander2014unified},
\emph{Students}~\cite{kronander2014unified},
\emph{ThrowingTowel}~\cite{kalantari_patch-based_2013}, and \emph{WavingHands}~\cite{kalantari_patch-based_2013}.  These test videos cover various
scenarios, including large camera and/or object motions, significant over-
and/or under-exposures. For the \emph{Fire}, \emph{ThrowingTowel} and \emph{WavingHands} datasets, the alternating frames are
available, while ground-truth radiance data for all the others exists. We
therefore generate the alternating frames of those sequences from the original
data by simulating the image acquisition procedure, {\it i.e.}, by alternating
the exposure times using the camera response function to obtain LDR frames.
Specifically, we use the camera response function
from~\cite{kalantari_patch-based_2013}, and set exposure times so that the
majority of the scenes are properly exposed in either the short or long
exposure. The exposure times for each video are listed in
Table~\ref{table:exposures}. In all experiments, $\alpha$ in~(\ref{eqn:obj}),
$\epsilon$ and $\lambda$ in~(\ref{eqn:objective}), and $\varepsilon$
in~(\ref{eqn:betaopt}) are fixed to $0.5$, $0.01$, and $0.1$, respectively. The
parameters $\beta$ and $\gamma$ in (\ref{eqn:obj}) are set online at each
iteration, {\it i.e.}, automatically updated at each iteration.  Specifically,
$\beta$ is set to $0.5\hat{\sigma}^2$, where $\hat{\sigma}$ is the standard
deviation of elements in $\cP_{\bS_k^\mathsf{c}}(\bD - \bB_k)$, and $\gamma =
\beta$. For \emph{Students}, however, as the intensity variations are small, we
set $\beta$ to be $0.01\hat{\sigma}^2$ and $\gamma=0.005\beta$. We vary
pyramidal level $L$ from $2$ to $4$ according to the magnitudes of object
motions to ensure a good compromise between computational cost and accuracy in
correspondences. We define the weighting matrix $\bW$ in~(\ref{eqn:obj}) as
\begin{equation}
    W_{i, j} =
    \begin{cases}
        w_s, & \text{if pixel $(i,j)$ is a spatial neighbor, }\\
        w_t, & \text{if pixel $(i,j)$ is a temporal neighbor, }\\
        0,   & \text{otherwise. }
    \end{cases}
\end{equation}
We set $w_s=0.2$ and $w_t=40$ for \emph{Students}
dataset, while $w_s=20$ and $w_t=20$ for all the others. We
experimentally found that MAP-HDR is robust to the choice
of parameters in terms of performance.

We compare the performance of the MAP-HDR algorithm
with three state-of-the-art HDR video synthesis methods: Kang {\it et al.}'s
algorithm~\cite{kang_high_2003}, Mangiat and Gibson's
algorithm~\cite{mangiat_high_2010,mangiat_spatially_2011}, and Kalantari {\it
    et al.}'s algorithm~\cite{kalantari_patch-based_2013}. For Kang {\it et
    al.}'s algorithm, we used our own implementation, while for Mangiat and
Gibson's and Kalantari {\it et al.}'s, we executed the codes provided by the
authors~\cite{code_Kalantari,code_Mangiat} with the optimal parameter settings to provide the best overall visual quality. To
print the synthesized HDR frames, we use the photographic tonemapping
technique~\cite{reinhard_photographic_2002} in all the experiments with the same parameter settings as in~\cite{kalantari_patch-based_2013}.

\subsection{Subjective Video Quality Assessment}
\label{ssec:subjective}

\begin{figure*}[!t]
	\centering
	\rotatebox[origin=c]{90}{\scriptsize Input}
	\subfloat {\includegraphics[width=3.2cm,valign=c]{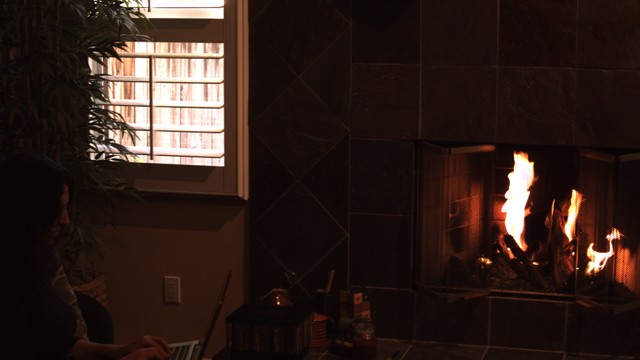}}\,\!
	\subfloat {\includegraphics[width=3.2cm,valign=c]{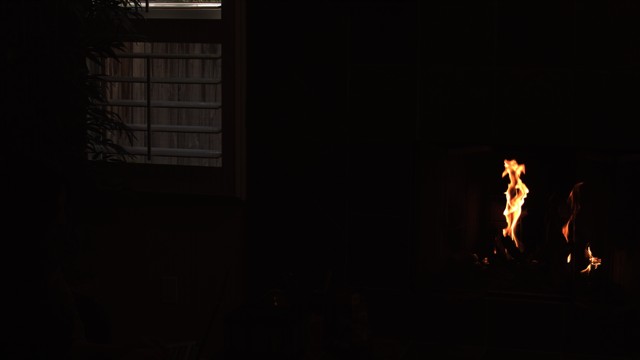}}\,\!
	\subfloat {\includegraphics[width=3.2cm,valign=c]{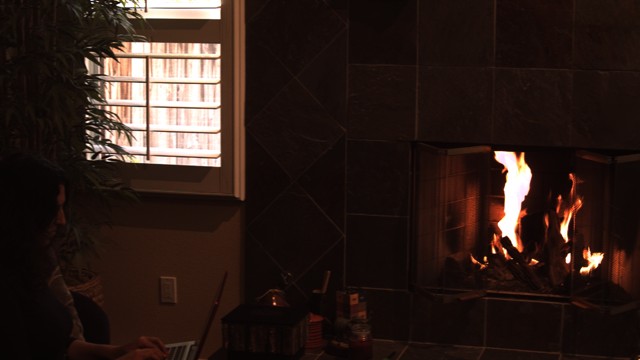}}\,\!
	\subfloat {\includegraphics[width=3.2cm,valign=c]{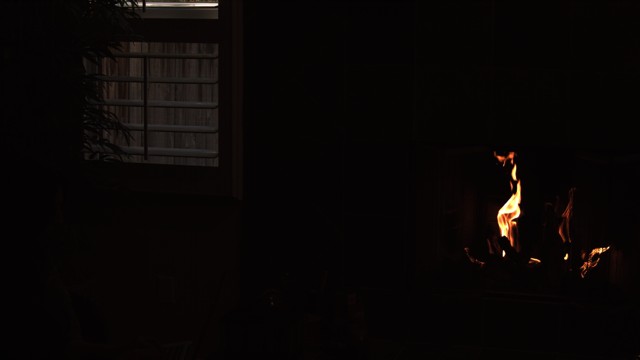}}\,\!
	\subfloat {\includegraphics[width=3.2cm,valign=c]{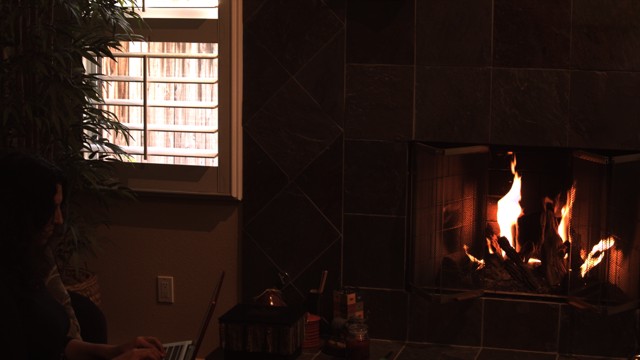}}\\
    \vspace{-2.5mm}

	\rotatebox[origin=c]{90}{\scriptsize Kang}
	\subfloat {\includegraphics[width=3.2cm,valign=c]{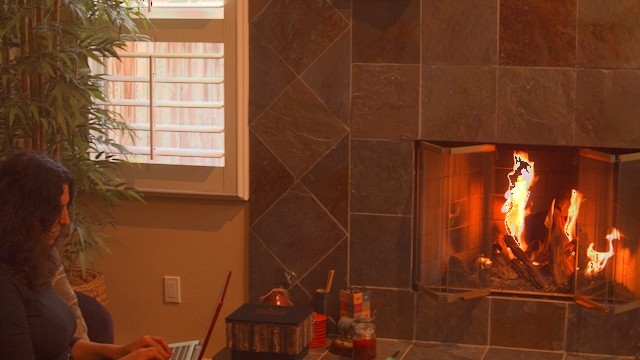}}\,\!
	\subfloat {\includegraphics[width=3.2cm,valign=c]{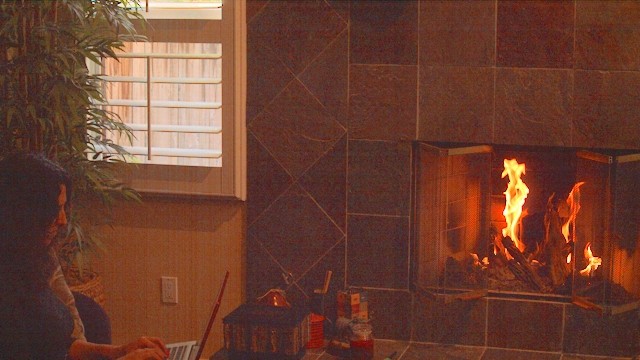}}\,\!
	\subfloat {\includegraphics[width=3.2cm,valign=c]{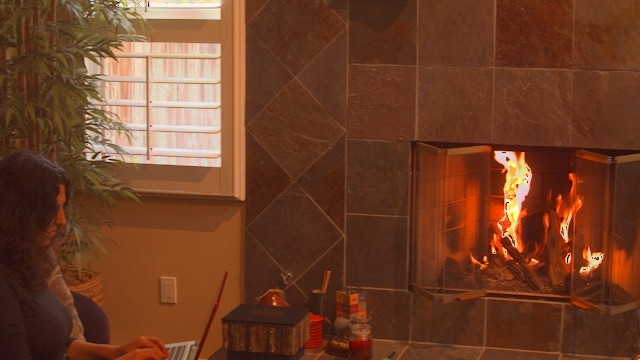}}\,\!
	\subfloat {\includegraphics[width=3.2cm,valign=c]{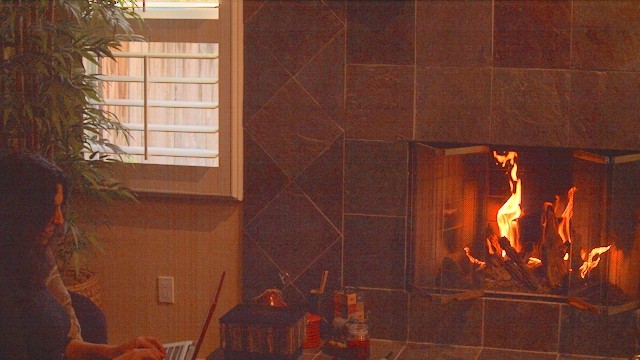}}\,\!
	\subfloat {\includegraphics[width=3.2cm,valign=c]{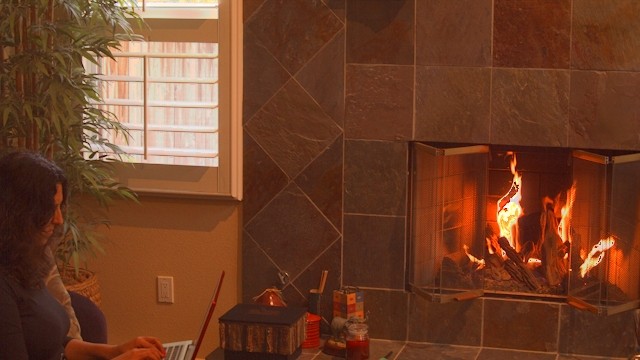}}\\
    \vspace{-2.5mm}

	\rotatebox[origin=c]{90}{\scriptsize Mangiat}
	\subfloat {\includegraphics[width=3.2cm,valign=c]{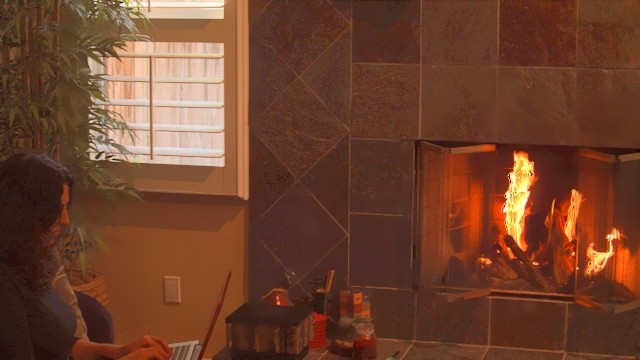}}\,\!
	\subfloat {\includegraphics[width=3.2cm,valign=c]{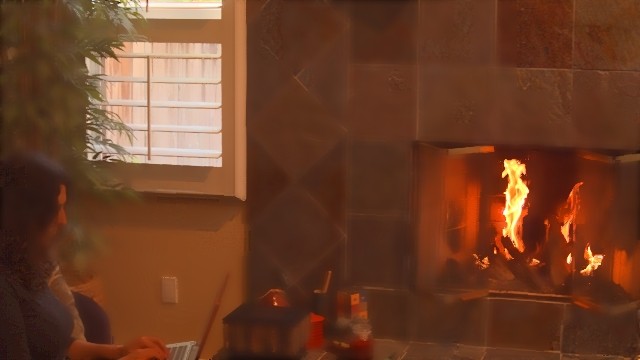}}\,\!
	\subfloat {\includegraphics[width=3.2cm,valign=c]{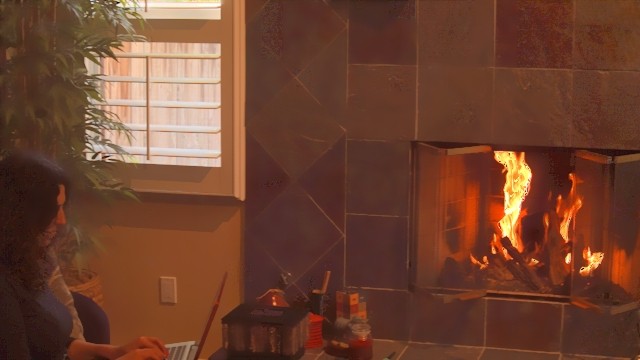}}\,\!
	\subfloat {\includegraphics[width=3.2cm,valign=c]{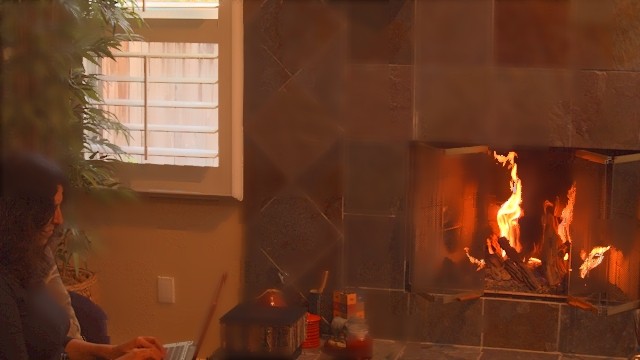}}\,\!
	\subfloat {\includegraphics[width=3.2cm,valign=c]{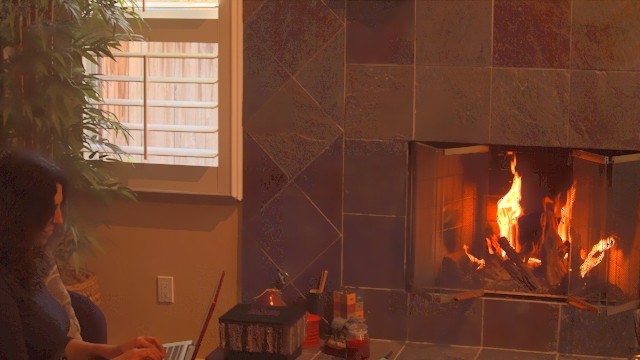}}\\
    \vspace{-2.5mm}

	\rotatebox[origin=c]{90}{\scriptsize Kalantari}\hspace{0.5mm}
	\subfloat {\includegraphics[width=3.2cm,valign=c]{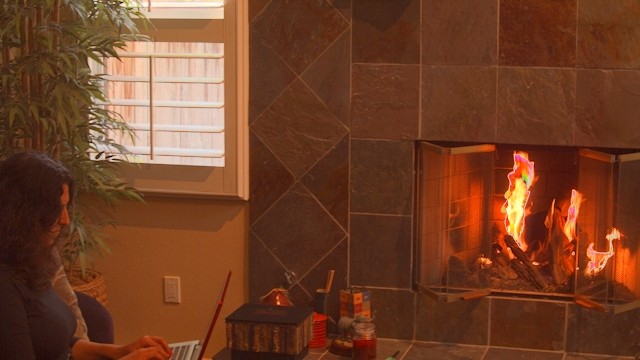}}\,\!
	\subfloat {\includegraphics[width=3.2cm,valign=c]{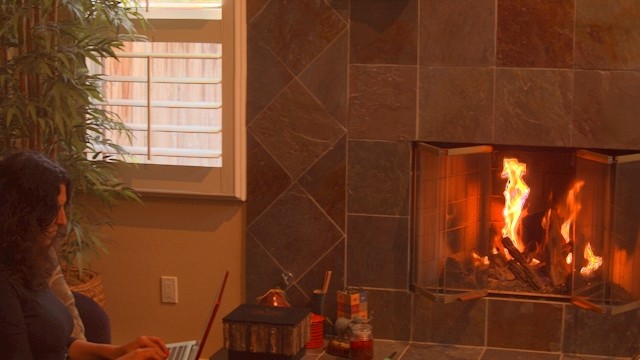}}\,\!
	\subfloat {\includegraphics[width=3.2cm,valign=c]{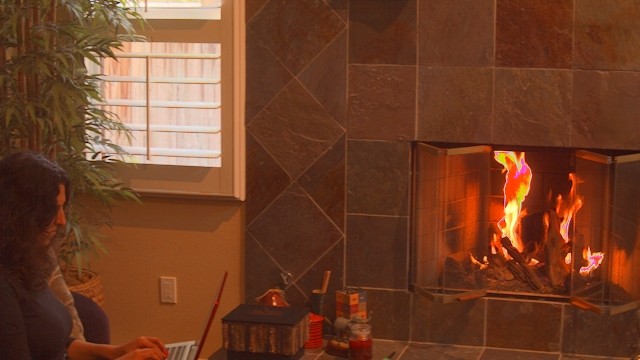}}\,\!
	\subfloat {\includegraphics[width=3.2cm,valign=c]{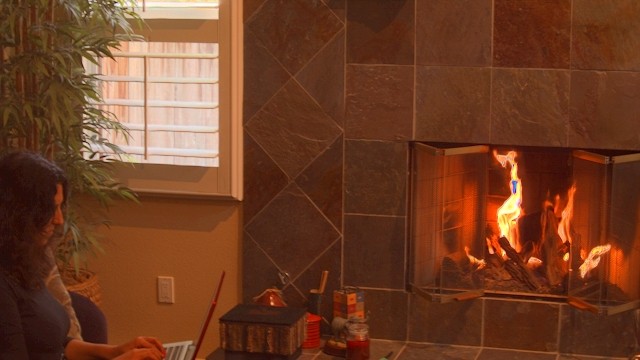}}\,\!
	\subfloat {\includegraphics[width=3.2cm,valign=c]{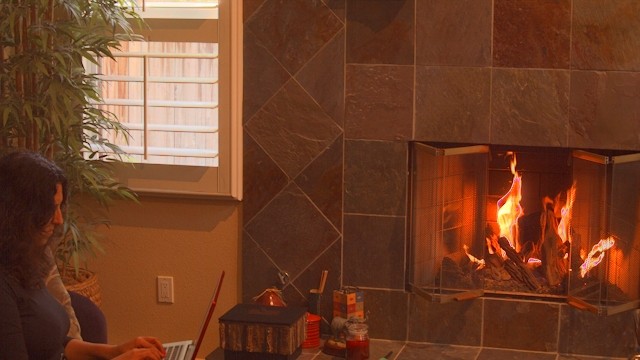}}\\
    \vspace{-2.5mm}

	\rotatebox[origin=c]{90}{\scriptsize MAP-HDR}\hspace{0.5mm}
	\subfloat {\includegraphics[width=3.2cm,valign=c]{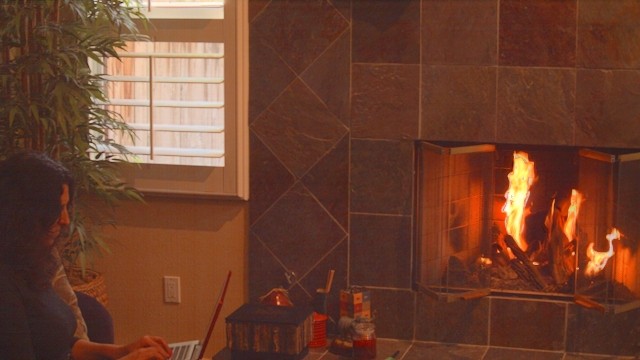}}\,\!
	\subfloat {\includegraphics[width=3.2cm,valign=c]{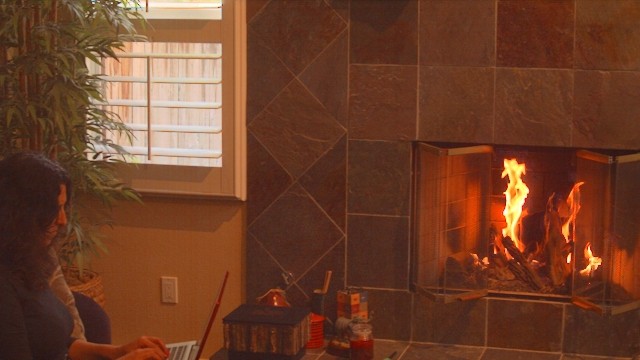}}\,\!
	\subfloat {\includegraphics[width=3.2cm,valign=c]{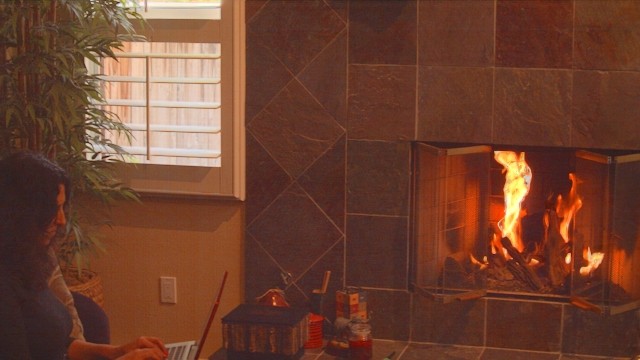}}\,\!
	\subfloat {\includegraphics[width=3.2cm,valign=c]{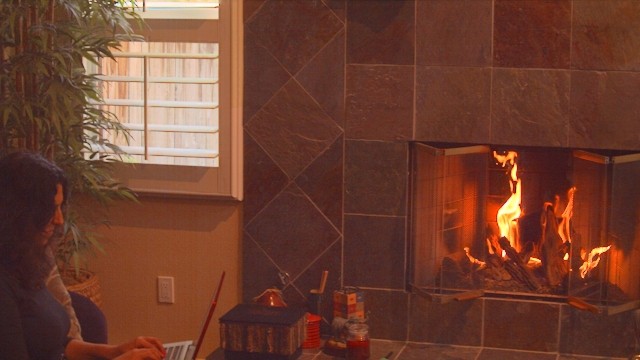}}\,\!
	\subfloat {\includegraphics[width=3.2cm,valign=c]{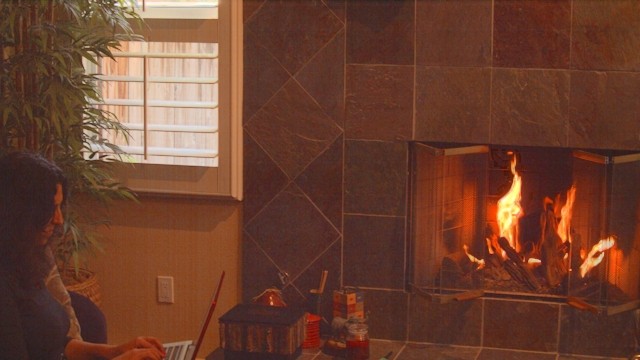}}\\
    \vspace{-2mm}

    \setcounter{subfigure}{0}
	\rotatebox[origin=c]{90}{\scriptsize Magnified}
	\subfloat [Input]{\includegraphics[trim={15cm 3.5cm 3.5cm 6cm},clip,width=3.2cm,valign=c]{figures/fire_005}}\,\!
	\subfloat [Kang {\it et al}.~\cite{kang_high_2003}]{\includegraphics[trim={15cm 3.5cm 3.5cm 6cm},clip,width=3.2cm,valign=c]{figures/Kang_fire_005}}\,\!
	\subfloat [Mangiat and Gibson~\cite{mangiat_spatially_2011}]{\includegraphics[trim={15cm 3.5cm 3.5cm 6cm},clip,width=3.2cm,valign=c]{figures/Mangiat_fire_005}}\,\!
	\subfloat [Kalantari {\it et al}.~\cite{kalantari_patch-based_2013}]{\includegraphics[trim={15cm 3.5cm 3.5cm 6cm},clip,width=3.2cm,valign=c]{figures/Kalantari_fire_005}}\,\!
	\subfloat [MAP-HDR]{\includegraphics[trim={15cm 3.5cm 3.5cm 6cm},clip,width=3.2cm,valign=c]{figures/rmhdr_fire_005}}\\

	\caption
	{
        HDR video synthesis results for the 4--8th frames of the \emph{Fire} sequence~\cite{kalantari_patch-based_2013}. The magnified parts of the 8th frame are shown in the last row
	}
	\label{fig:fire}
\end{figure*}

\begin{figure*}
	\centering
	\rotatebox[origin=c]{90}{\scriptsize Input}
	\subfloat {\includegraphics[width=3.2cm,valign=c]{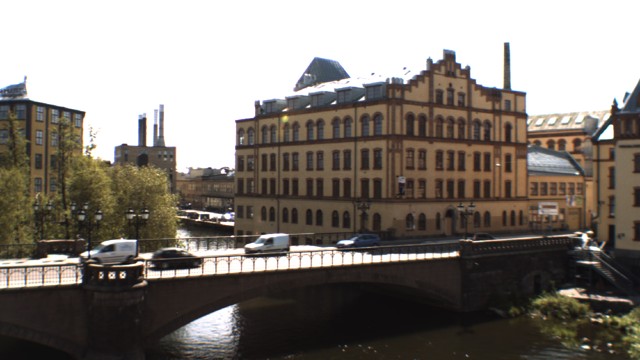}}\,\!
	\subfloat {\includegraphics[width=3.2cm,valign=c]{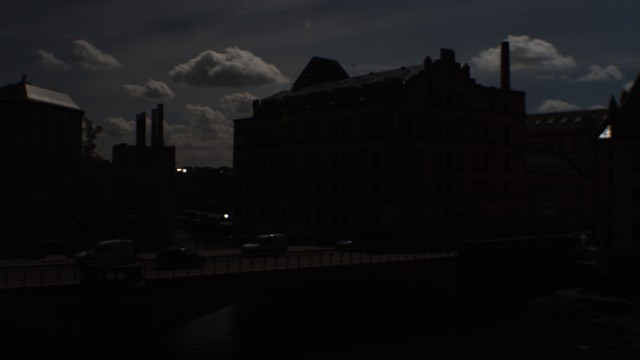}}\,\!
	\subfloat {\includegraphics[width=3.2cm,valign=c]{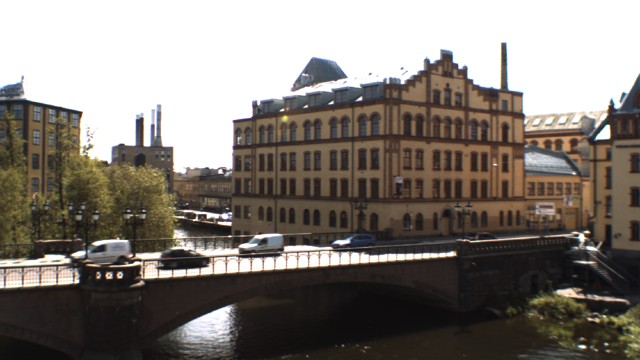}}\,\!
	\subfloat {\includegraphics[width=3.2cm,valign=c]{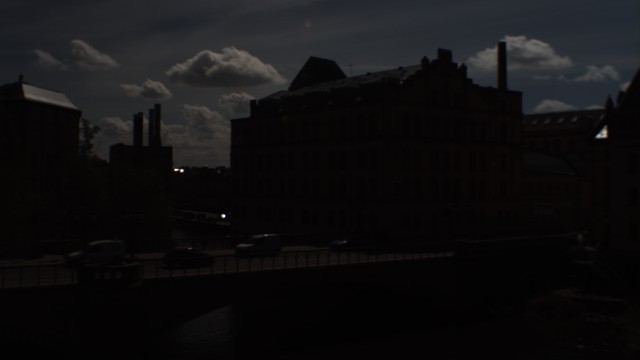}}\,\!
	\subfloat {\includegraphics[width=3.2cm,valign=c]{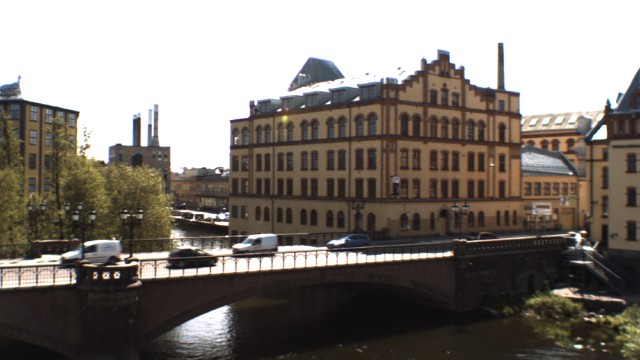}}\\
    \vspace{-2.5mm}

	\rotatebox[origin=c]{90}{\scriptsize Kang}
	\subfloat {\includegraphics[width=3.2cm,valign=c]{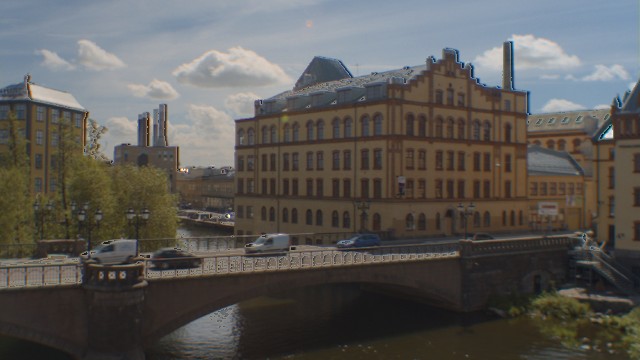}}\,\!
	\subfloat {\includegraphics[width=3.2cm,valign=c]{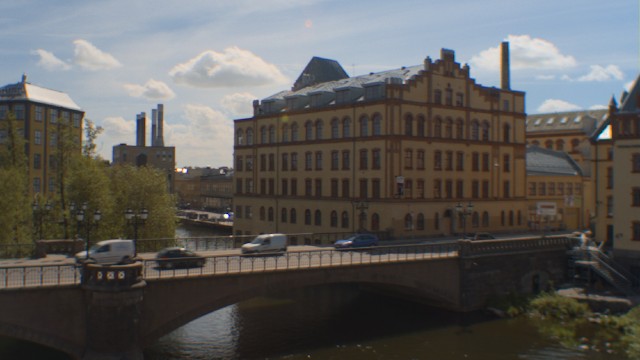}}\,\!
	\subfloat {\includegraphics[width=3.2cm,valign=c]{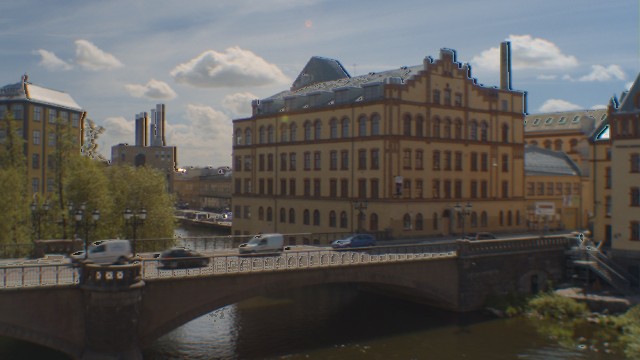}}\,\!
	\subfloat {\includegraphics[width=3.2cm,valign=c]{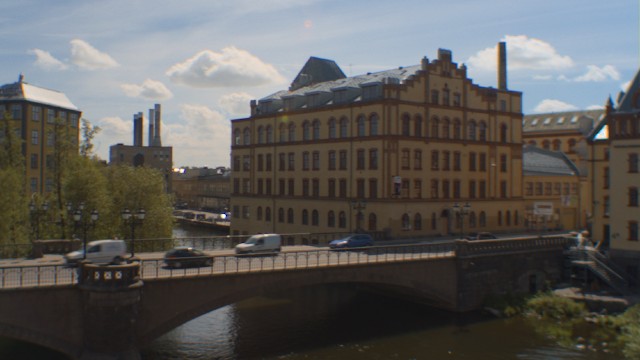}}\,\!
	\subfloat {\includegraphics[width=3.2cm,valign=c]{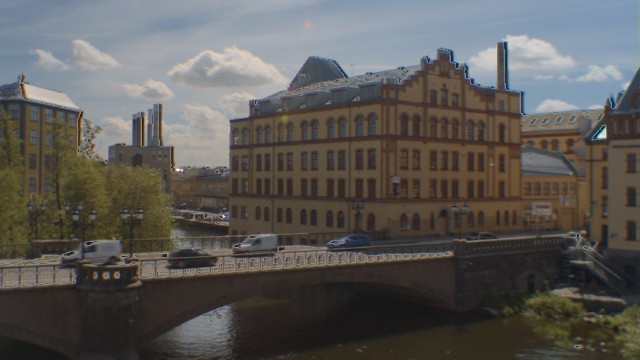}}\\
    \vspace{-2.5mm}

	\rotatebox[origin=c]{90}{\scriptsize Mangiat}
	\subfloat {\includegraphics[width=3.2cm,valign=c]{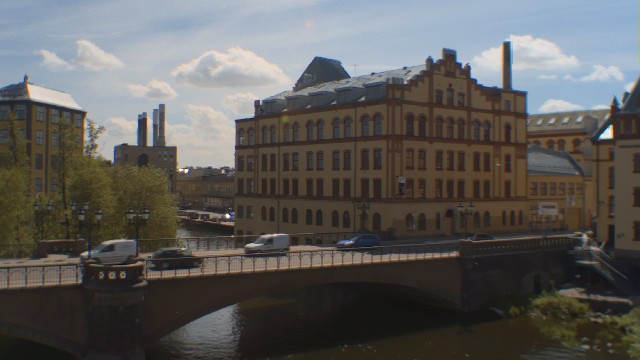}}\,\!
	\subfloat {\includegraphics[width=3.2cm,valign=c]{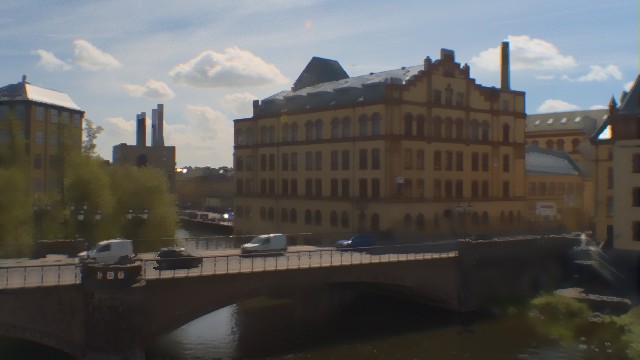}}\,\!
	\subfloat {\includegraphics[width=3.2cm,valign=c]{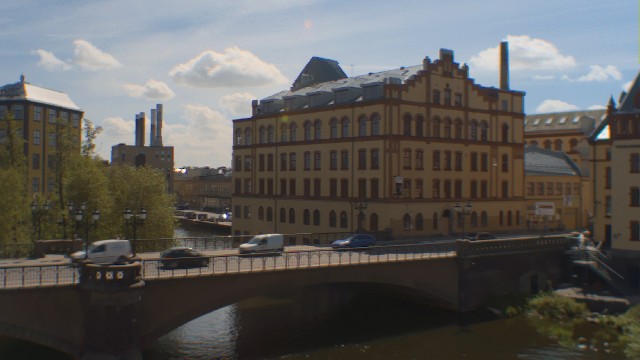}}\,\!
	\subfloat {\includegraphics[width=3.2cm,valign=c]{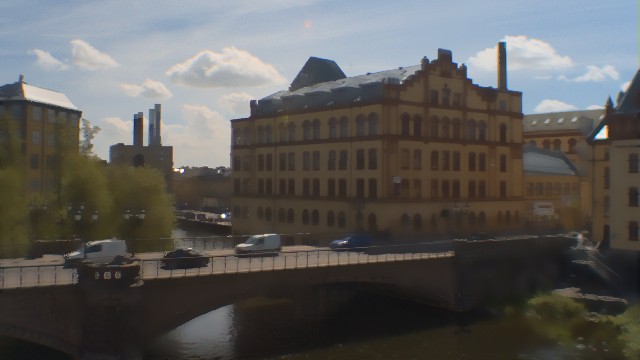}}\,\!
	\subfloat {\includegraphics[width=3.2cm,valign=c]{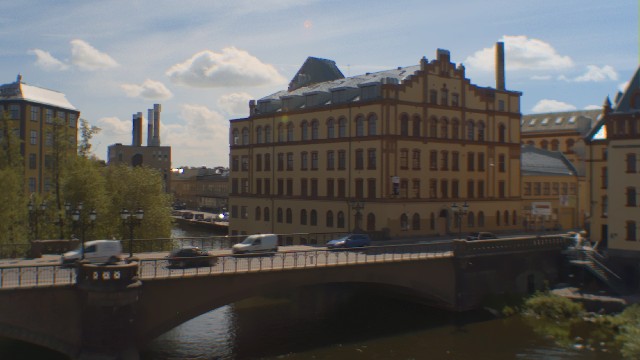}}\\
    \vspace{-2.5mm}

	\rotatebox[origin=c]{90}{\scriptsize Kalantari}\hspace{0.5mm}
	\subfloat {\includegraphics[width=3.2cm,valign=c]{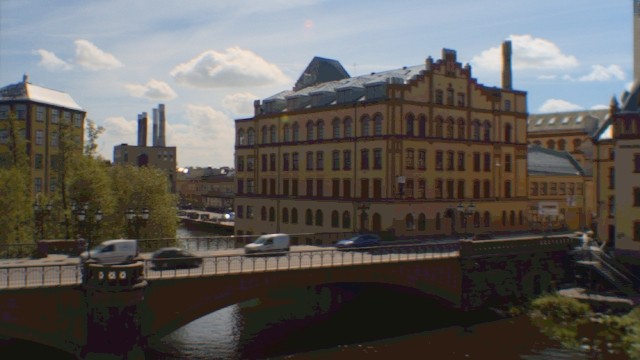}}\,\!
	\subfloat {\includegraphics[width=3.2cm,valign=c]{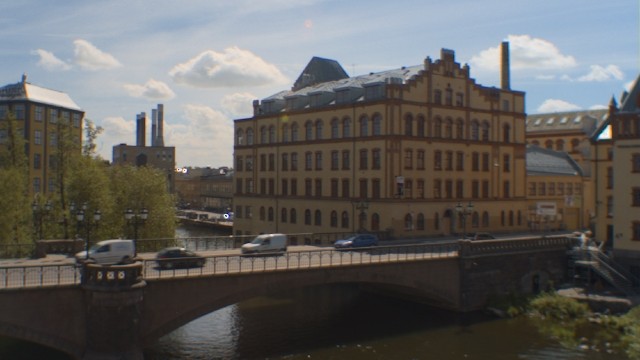}}\,\!
	\subfloat {\includegraphics[width=3.2cm,valign=c]{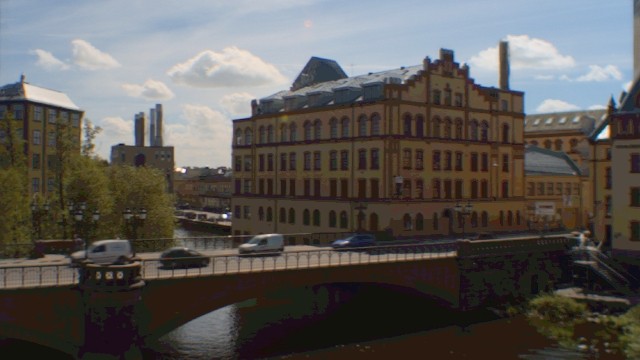}}\,\!
	\subfloat {\includegraphics[width=3.2cm,valign=c]{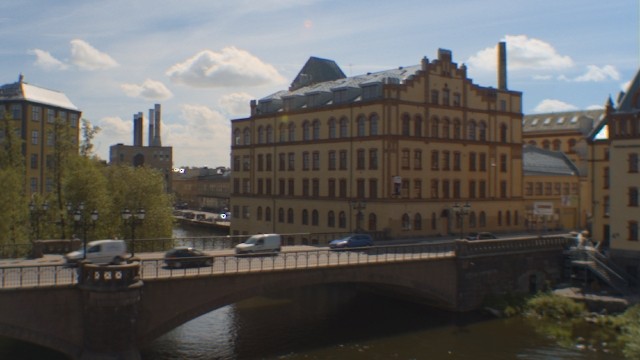}}\,\!
	\subfloat {\includegraphics[width=3.2cm,valign=c]{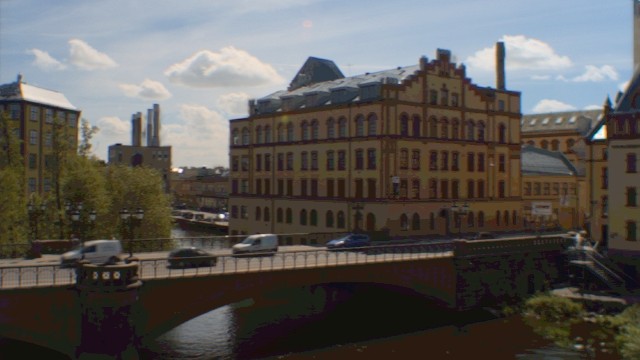}}\\
    \vspace{-2.5mm}

    \rotatebox[origin=c]{90}{\scriptsize MAP-HDR}\hspace{0.5mm}
    \subfloat {\includegraphics[width=3.2cm,valign=c]{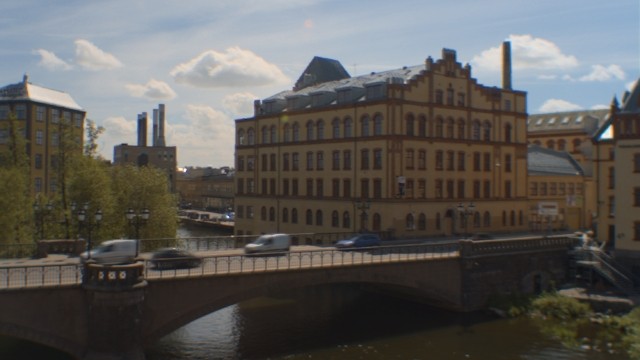}}\,\!
    \subfloat {\includegraphics[width=3.2cm,valign=c]{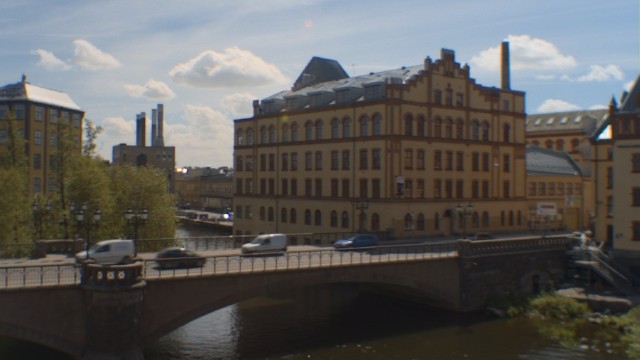}}\,\!
    \subfloat {\includegraphics[width=3.2cm,valign=c]{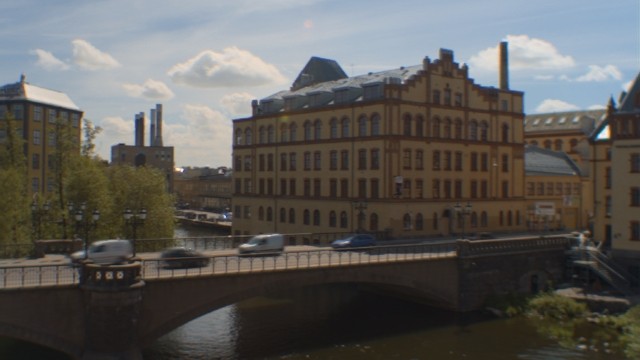}}\,\!
    \subfloat {\includegraphics[width=3.2cm,valign=c]{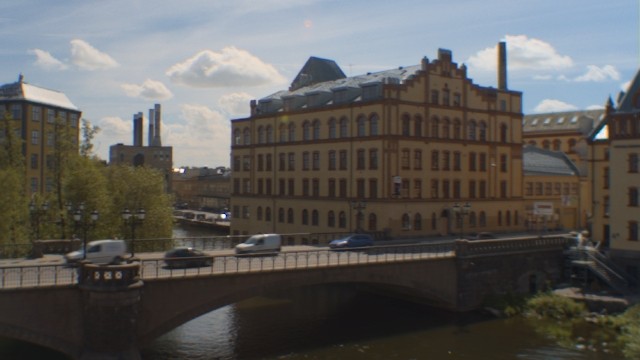}}\,\!
    \subfloat {\includegraphics[width=3.2cm,valign=c]{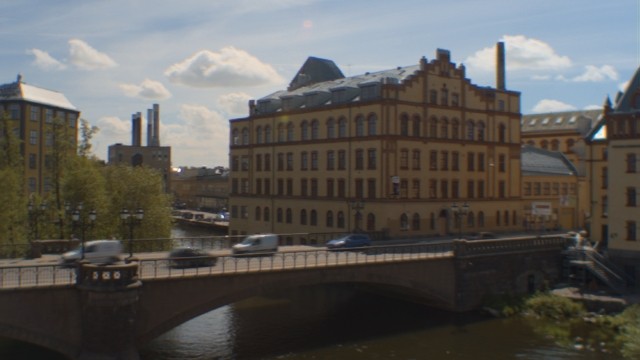}}\\
    \vspace{-2mm}


    \setcounter{subfigure}{0}
    \rotatebox[origin=c]{90}{\scriptsize Magnified}
    \subfloat [Input]{\includegraphics[trim={140px 30px 340px 220px},clip,width=3.2cm,valign=c]{figures/bridge2_005}}\,\!
    \subfloat [Kang {\it et al}.~\cite{kang_high_2003}]{\includegraphics[trim={140px 30px 340px 220px},clip,width=3.2cm,valign=c]{figures/Kang_bridge2_005}}\,\!
    \subfloat [Mangiat and Gibson]{\includegraphics[trim={140px 30px 340px 220px},clip,width=3.2cm,valign=c]{figures/Mangiat_bridge2_005}}\,\!
    \subfloat [Kalantari {\it et al}.~\cite{kalantari_patch-based_2013}]{\includegraphics[trim={140px 30px 340px 220px},clip,width=3.2cm,valign=c]{figures/Kalantari_bridge2_005}}\,\!
    \subfloat [MAP-HDR]{\includegraphics[trim={140px 30px 340px 220px},clip,width=3.2cm,valign=c]{figures/rmhdr_bridge2_005}}\\

    \caption{
        HDR video synthesis results for the 3--7th frames of the \emph{Bridge2} sequence~\cite{kronander2014unified}. The magnified parts of the 3rd frame are shown in the last row.
    }
    \label{fig:bridge2}
\end{figure*}

Fig.~\ref{fig:fire} shows the 4--8th frames of the \emph{Fire} sequence and the
results synthesized by each algorithm. Kang {\it et al.}'s
algorithm~\cite{kang_high_2003} in the second row provides artifacts inside the
flames. This is because of its mechanism of compositing the unidirectionally
warped frames with bidirectionally warped frames, which may have inconsistent
content under mis-registration scenarios.  Mangiat and Gibson's
algorithm~\cite{mangiat_spatially_2011} in the third row performs better than
Kang {\it et al.}'s, but it still produces artifacts in the flame, as a
consequence of mis-registration. For example, it severely blurs the facade of the fireplace of 5th
and 7th frames, since its HDR filtering may produce blurred results in the case
of mis-registration. Kalantari {\it et al.}'s
algorithm~\cite{kalantari_patch-based_2013} in the fourth row preserves the
shape of flames but introduces color artifacts due to the failure of
correspondences estimation. On the contrary, the MAP-HDR algorithm in the
fifth row most faithfully renders the video frames with only mild artifacts,
which are not even noticeable in the resulting videos.

Fig.~\ref{fig:bridge2} shows the 3--7th frames of the \emph{Bridge2} sequence
with synthesized results. Kang {\it et al.}'s algorithm works well for the
frames with low exposures, whereas it produces severe artifacts for the
others. Frames with low exposures usually have few saturated regions, and in
such frames Kang {\it et al.}'s algorithm maintains most of the original
content. However, when a frame has rich saturation, severe artifacts around edges are
brought in from bidirectionally warped frames, especially in cases of
mis-registrations. Mangiat and Gibson's algorithm alleviates the artifacts via
HDR filtering at the cost of blurring artifacts, {\it e.g.}, on the facets of the buildings in the 4th and 6th frames. Kalantari {\it et al.}'s
method distorts the edges of buildings, and also distorts the color in the
facet of the bridge. In contrast, we can see that the MAP-HDR algorithm
produces higher-quality HDR frames without noticeable artifacts.

\begin{figure*}[!t]
    \centering
    \subfloat {\includegraphics[trim={10px 250px 580px 70px},clip,width=3.4cm]{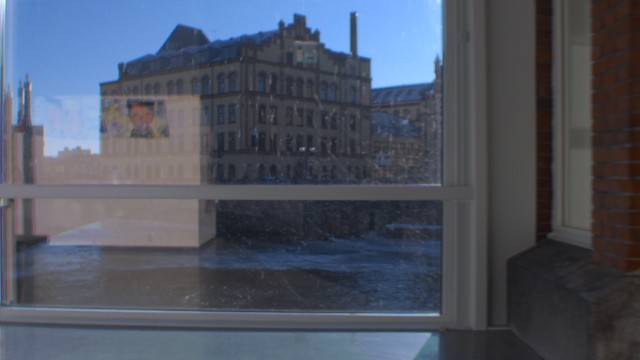}}\,\!
    \subfloat {\includegraphics[trim={10px 250px 580px 70px},clip,width=3.4cm]{figures/groundtruth_hallway2_001}}\,\!
    \subfloat {\includegraphics[trim={10px 250px 580px 70px},clip,width=3.4cm]{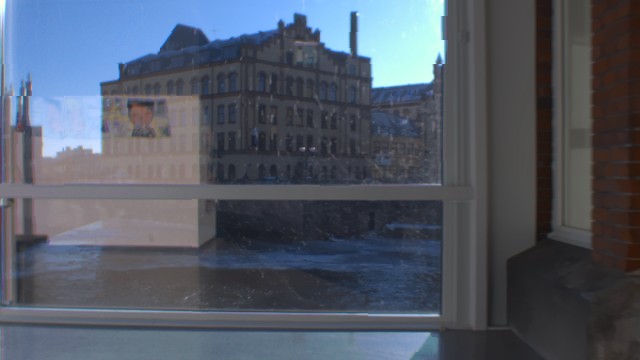}}\,\!
    \subfloat {\includegraphics[trim={10px 250px 580px 70px},clip,width=3.4cm]{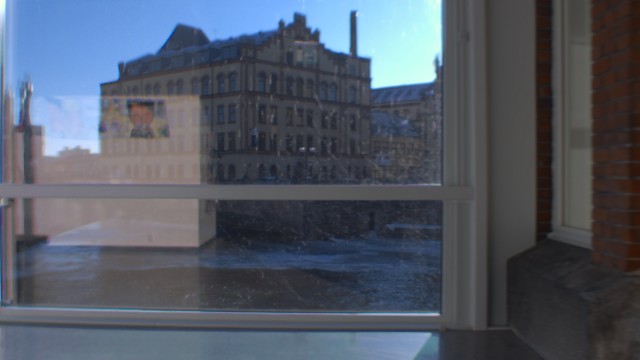}}\,\!
    \subfloat {\includegraphics[trim={10px 250px 580px 70px},clip,width=3.4cm]{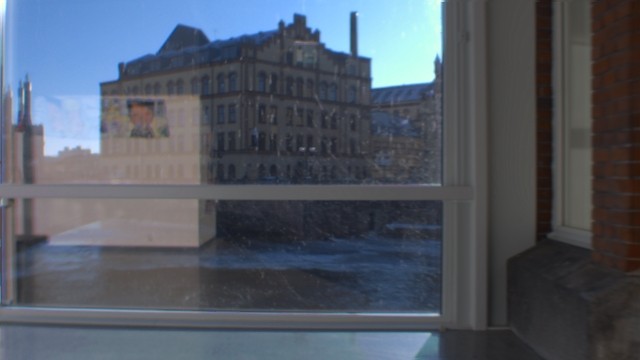}}\\

    \vspace{-2.5mm}
    \setcounter{subfigure}{0}
    \subfloat[Ground truth] {\includegraphics[trim={10px 250px 580px 70px},clip,width=3.4cm]{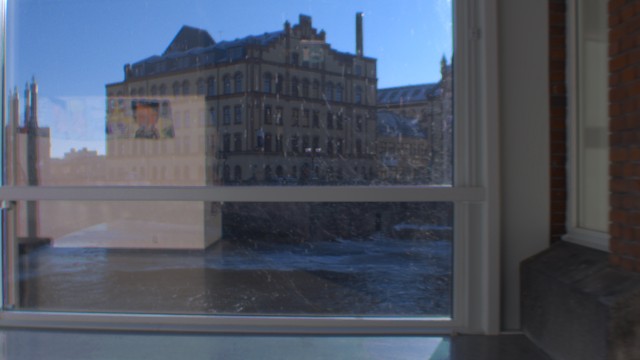}}\,\!
    \subfloat[Kang {\it et al.}~\cite{kang_high_2003}] {\includegraphics[trim={10px 250px 580px 70px},clip,width=3.4cm]{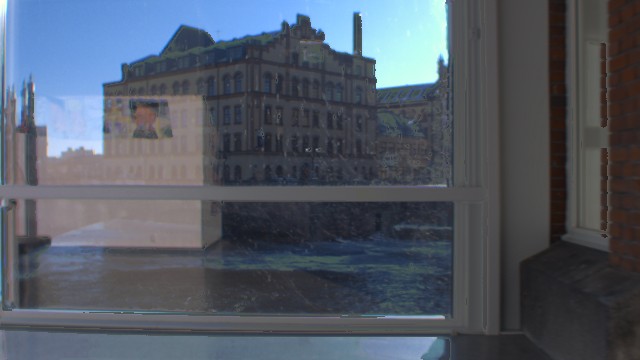}}\,\!
    \subfloat[Mangiat and Gibson~\cite{mangiat_spatially_2011}] {\includegraphics[trim={10px 250px 580px 70px},clip,width=3.4cm]{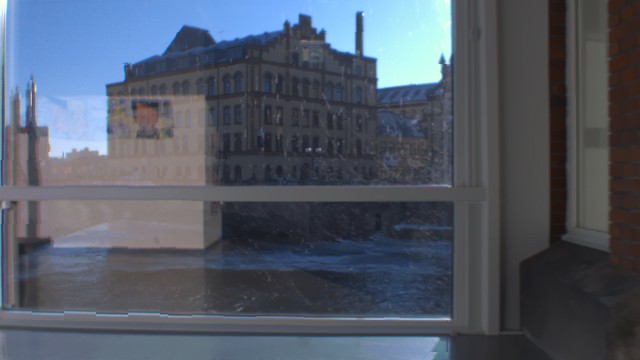}}\,\!
    \subfloat[Kalantari {\it et al.}~\cite{kalantari_patch-based_2013}] {\includegraphics[trim={10px 250px 580px 70px},clip,width=3.4cm]{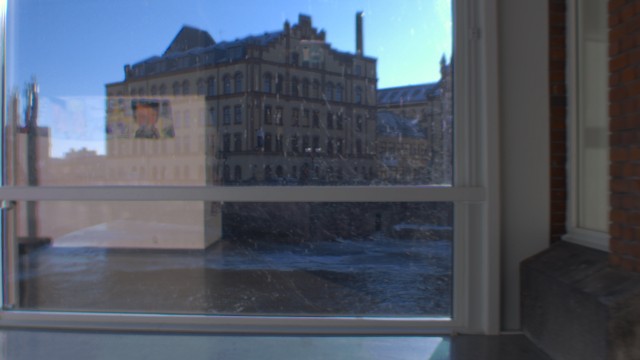}}\,\!
    \subfloat[MAP-HDR] {\includegraphics[trim={10px 250px 580px 70px},clip,width=3.4cm]{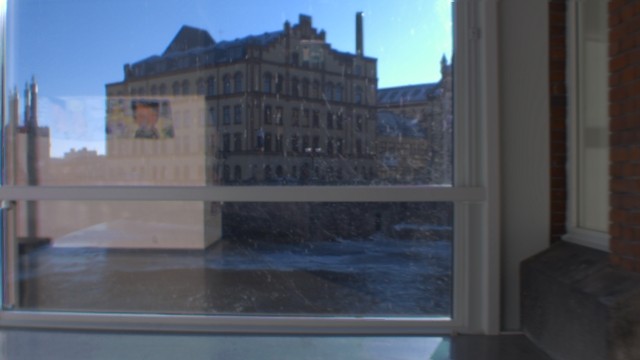}}\\

    \caption
    {
        Synthesized results of the 142th frame (top row) and 146th frame (bottom row) on the
        \emph{Hallway2} sequence~\cite{kronander2014unified}.
    }
    \label{fig:hallway2}
\end{figure*}

\begin{figure*}[!t]
    \centering
    \subfloat {\includegraphics[trim={20px 80px 200px 80px},clip,width=3.4cm]{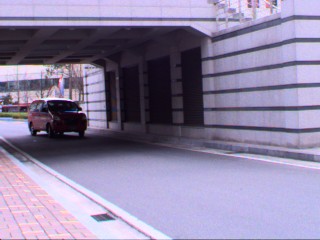}}\,\!
    \subfloat {\includegraphics[trim={20px 80px 200px 80px},clip,width=3.4cm]{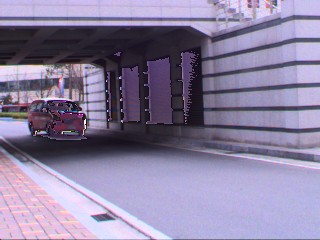}}\,\!
    \subfloat {\includegraphics[trim={20px 80px 200px 80px},clip,width=3.4cm]{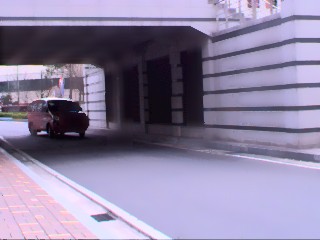}}\,\!
    \subfloat {\includegraphics[trim={20px 80px 200px 80px},clip,width=3.4cm]{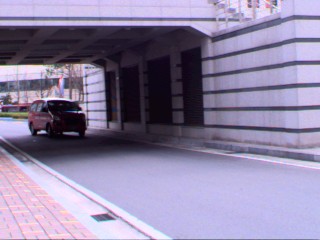}}\,\!
    \subfloat {\includegraphics[trim={20px 80px 200px 80px},clip,width=3.4cm]{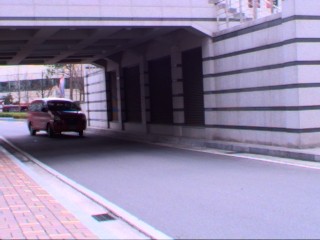}}\\

    \vspace{-2.5mm}
    \setcounter{subfigure}{0}
    \subfloat [Ground truth] {\includegraphics[trim={20px 80px 200px 80px},clip,width=3.4cm,valign=c]{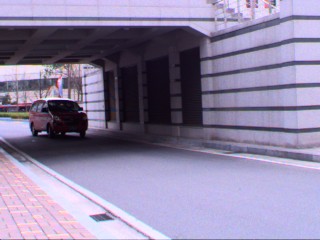}}\,\!
    \subfloat [Kang {\it et al.}~\cite{kang_high_2003}] {\includegraphics[trim={20px 80px 200px 80px},clip,width=3.4cm,valign=c]{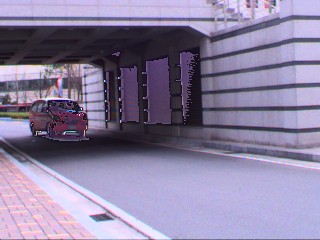}}\,\!
    \subfloat [Mangiat and Gibson~\cite{mangiat_spatially_2011}] {\includegraphics[trim={20px 80px 200px 80px},clip,width=3.4cm,valign=c]{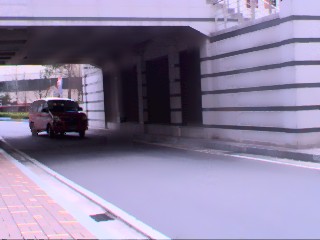}}\,\!
    \subfloat [Kalantari {\it et al.}~\cite{kalantari_patch-based_2013}] {\includegraphics[trim={20px 80px 200px 80px},clip,width=3.4cm,valign=c]{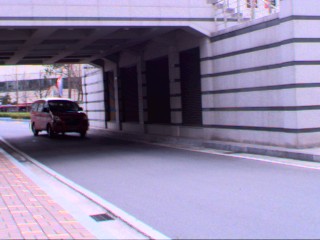}}\,\!
    \subfloat [MAP-HDR] {\includegraphics[trim={20px 80px 200px 80px},clip,width=3.4cm,valign=c]{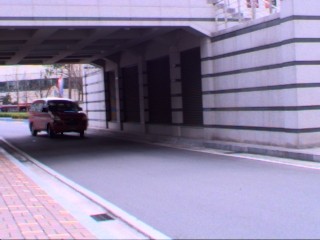}}\\

    \caption
    {
        Synthesized results of the 51th frame (top row) and 53th frame (bottom row) on the
        \emph{ParkingLot} sequence~\cite{lee_rate-distortion_2012}.
    }
    \label{fig:parkinglot}
\end{figure*}

\begin{figure*}[!t]
    \centering
    \subfloat {\includegraphics[trim={100px 0px 320px 200px},clip,width=3.4cm]{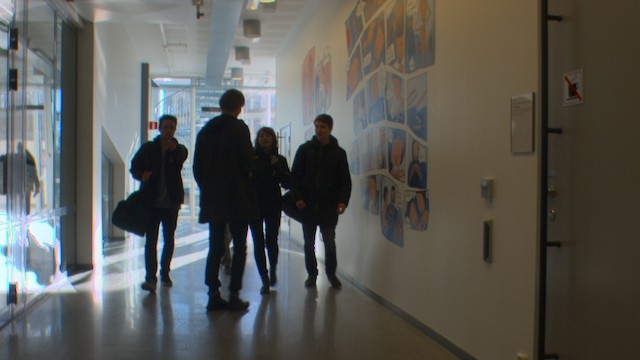}}\,\!
    \subfloat {\includegraphics[trim={100px 0px 320px 200px},clip,width=3.4cm]{figures/groundtruth_students_001}}\,\!
    \subfloat {\includegraphics[trim={100px 0px 320px 200px},clip,width=3.4cm]{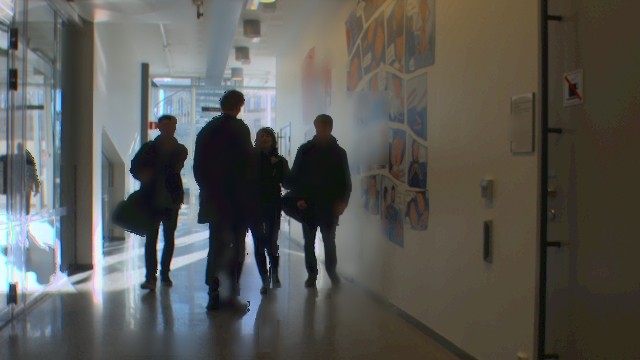}}\,\!
    \subfloat {\includegraphics[trim={100px 0px 320px 200px},clip,width=3.4cm]{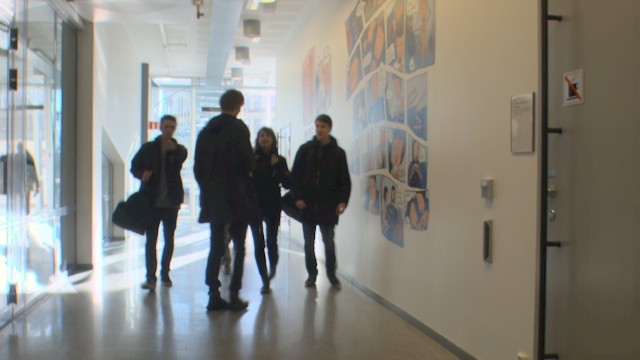}}\,\!
    \subfloat {\includegraphics[trim={100px 0px 320px 200px},clip,width=3.4cm]{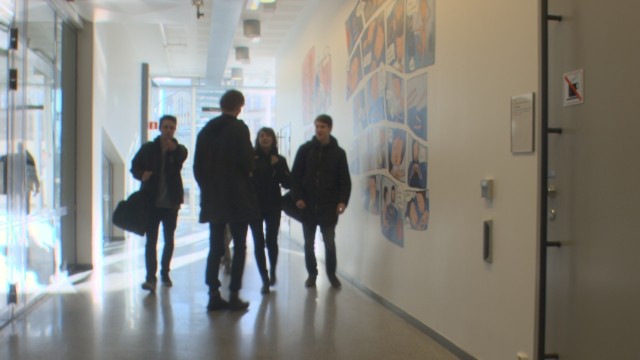}}\\

    \vspace{-2.5mm}
    \setcounter{subfigure}{0}
    \subfloat[Ground truth] {\includegraphics[trim={100px 0px 320px 200px},clip,width=3.4cm]{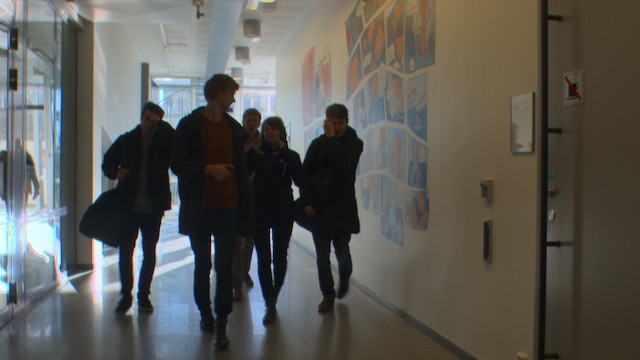}}\,\!
    \subfloat[Kang {\it et al.}~\cite{kang_high_2003}] {\includegraphics[trim={100px 0px 320px 200px},clip,width=3.4cm]{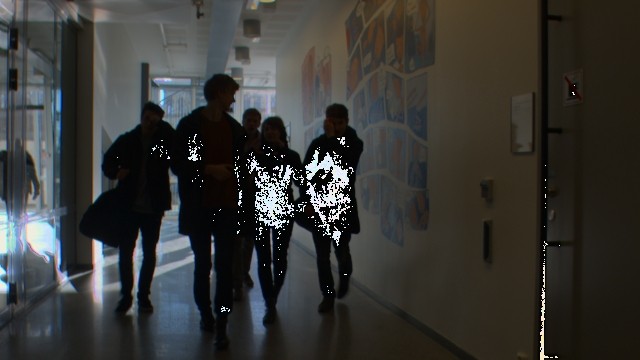}}\,\!
    \subfloat[Mangiat and Gibson~\cite{mangiat_spatially_2011}] {\includegraphics[trim={100px 0px 320px 200px},clip,width=3.4cm]{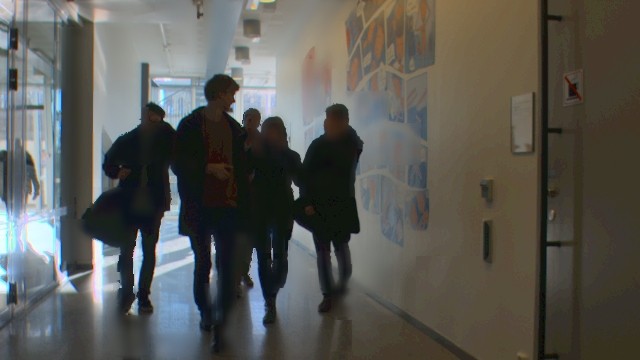}}\,\!
    \subfloat[Kalantari {\it et al.}~\cite{kalantari_patch-based_2013}] {\includegraphics[trim={100px 0px 320px 200px},clip,width=3.4cm]{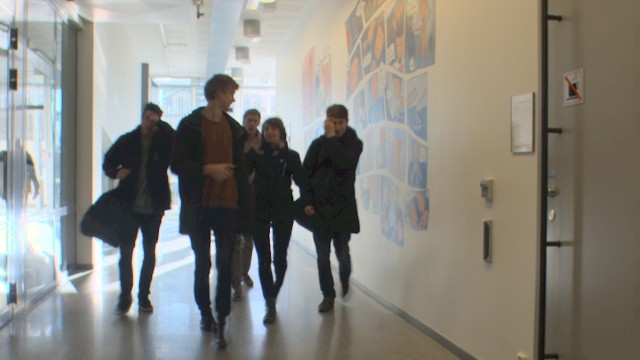}}\,\!
    \subfloat[MAP-HDR] {\includegraphics[trim={100px 0px 320px 200px},clip,width=3.4cm]{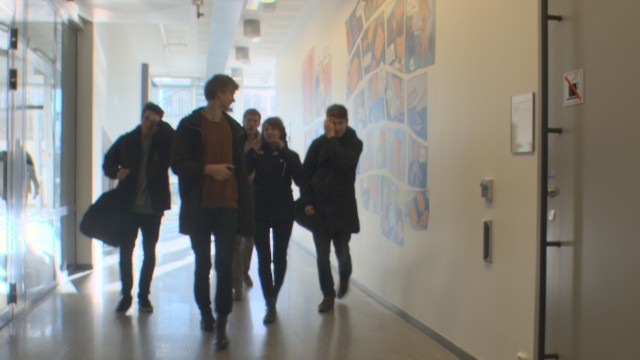}}\\

    \caption{
        Synthesized results for the 11th and 37th frames of the \emph{Students} sequence~\cite{kronander2014unified}.
    }
    \label{fig:students}
\end{figure*}

Figs.~\ref{fig:hallway2}--\ref{fig:students} compare magnified parts of
the results on the \emph{Hallway2}, \emph{ParkingLot}, and \emph{Students} sequences,
respectively. In Fig.~\ref{fig:hallway2}, Kang {\it et al.}'s
method~\cite{kang_high_2003} does not noticeably distort the objects in the
146th frame, while it blends the tower in the 142th frame. Moreover, it
introduces noticeable color artifacts. This is because of mis-registration in the
bidirectionally warped frames and color inconsistencies in the unidirectionally
warped frames. Mangiat and Gibson's algorithm~\cite{mangiat_spatially_2011} and
Kalantari {\it et al}.'s algorithm~\cite{kalantari_patch-based_2013} both
introduce visible artifacts. In particular, they both distort the tower of the
building. This is due to the failure of optical flow in saturated regions. In
the original LDR frames with long exposures, the tower belongs to saturated
regions, and optical flow fails to accurately estimate its correspondences.
Therefore, both Mangiat and Gibson's and Kalantari {\it et al.}'s algorithms
fail to provide high-quality frames, since they both rely on the accuracy of
optical flow. In contrast, the MAP-HDR algorithm well preserves the original
shape of the tower due to the robustness of the proposed kernel regression
scheme discussed in Section~\ref{ssec:lakr}.

In Fig.~\ref{fig:parkinglot}, Kang {\it et al.}'s algorithm preserves the
shape of the van but introduces severe color artifacts in the front of the van because of the
inconsistencies between bidirectionally warped frames and unidirectionally
warped ones. Mangiat and Gibson's algorithm provides ghosting artifacts near
the tail of the van and erodes the text on the van in the 51th frame due to mis-registration.
Kalantari {\it et al.}'s algorithm distorts the shape of the van, since its
patch-based search fails to fill the saturated region faithfully with content
from neighboring frames. In comparison, the MAP-HDR algorithm yields HDR
frames without object distortions or ghosting artifacts, while its mild
blurring effect is generally unnoticeable in the resulting video.

Fig.~\ref{fig:students} compares the synthesized results on the \emph{Students} sequence, which contains large object motions. Kang {\it et al.}'s algorithm~\cite{kang_high_2003} provides severe artifacts, {\it e.g.}, white pixels on a student's body. Mangiat and Gibson's algorithm~\cite{mangiat_spatially_2011} yields clearly visible blurring artifacts, especially around students' legs and feet, where there are complex and large object motions. Kalantari {\it et al.}'s algorithm~\cite{kalantari_patch-based_2013} distorts the shape of the moving objects, {\it e.g.}, the students' legs in the 11th frame are shrunk. Moreover, these distortions result in temporal incoherences.\footnote{Please see the resulting videos in the supplementary material as well as on our project website.} In contrast, the MAP-HDR algorithm produces higher-quality results with significantly less artifact both spatially and temporally by employing the multiscale locally adaptive kernel regression.

\begin{figure*}[t]
    \centering

    \subfloat  {\includegraphics[width=3.5cm]{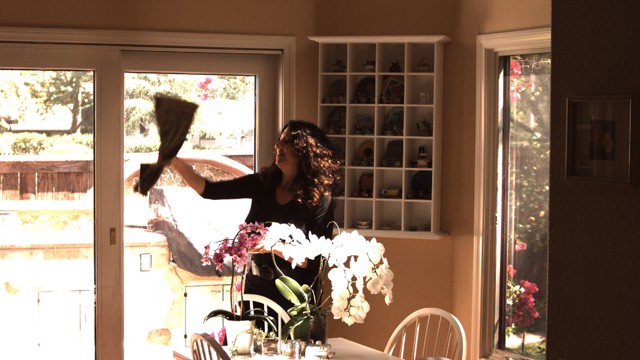}}\,\!
    \subfloat  {\includegraphics[width=3.5cm]{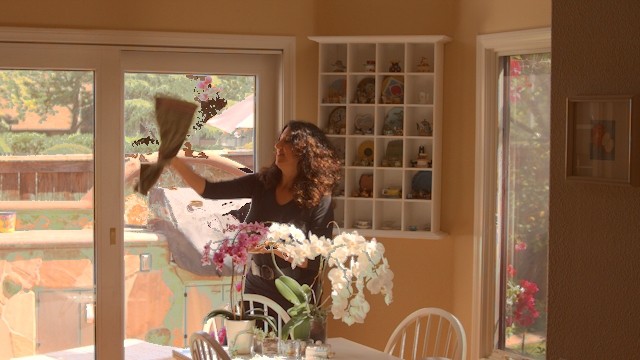}}\,\!
    \subfloat  {\includegraphics[width=3.5cm]{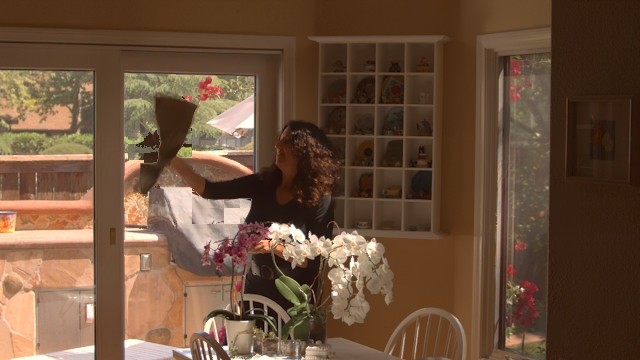}}\,\!
    \subfloat  {\includegraphics[width=3.5cm]{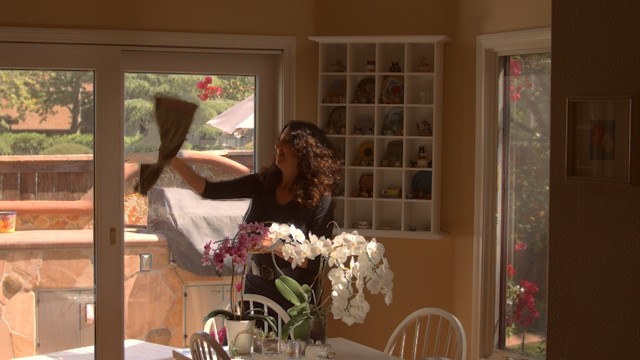}}\,\!
    \subfloat  {\includegraphics[width=3.5cm]{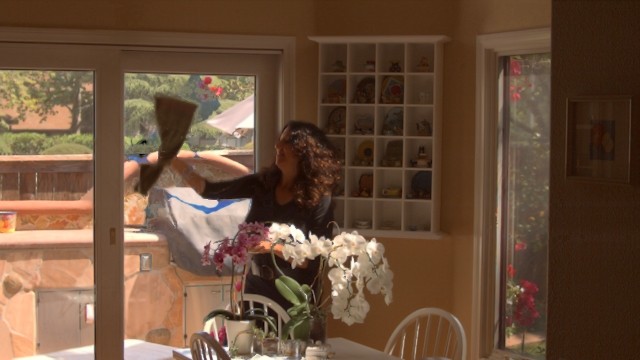}}\\

    \vspace{-2.5mm}
    \setcounter{subfigure}{0}
    \subfloat [Input] {\includegraphics[width=3.5cm]{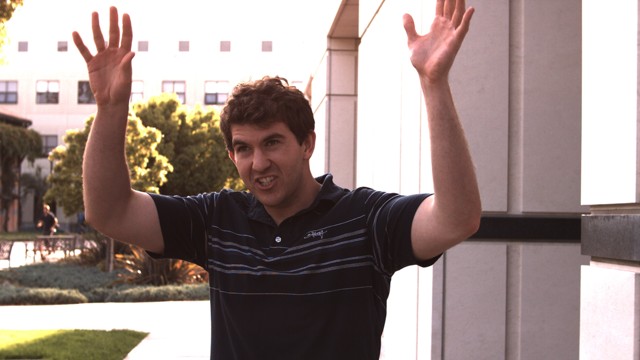}}\,\!
    \subfloat [Kang {\it et al.}~\cite{kang_high_2003}] {\includegraphics[width=3.5cm]{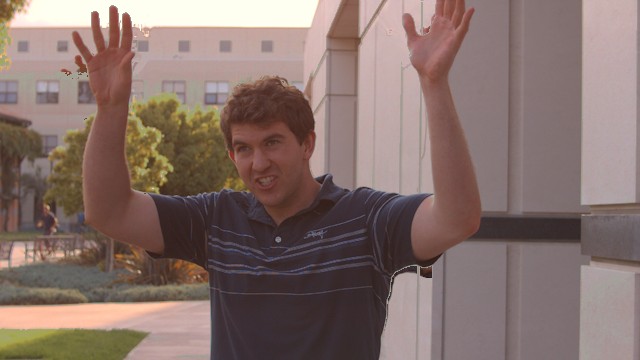}}\,\!
    \subfloat [Mangiat and Gibson~\cite{mangiat_spatially_2011}] {\includegraphics[width=3.5cm]{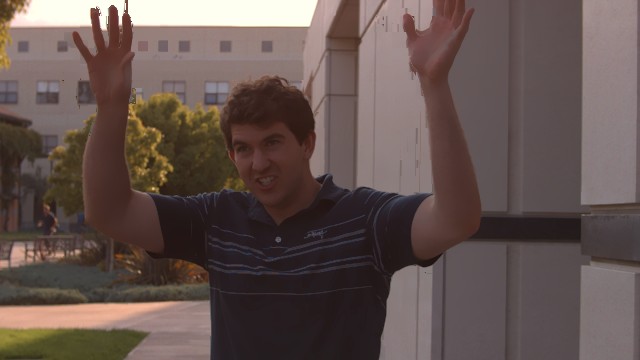}}\,\!
    \subfloat [Kalantari {\it et al.}~\cite{kalantari_patch-based_2013}] {\includegraphics[width=3.5cm]{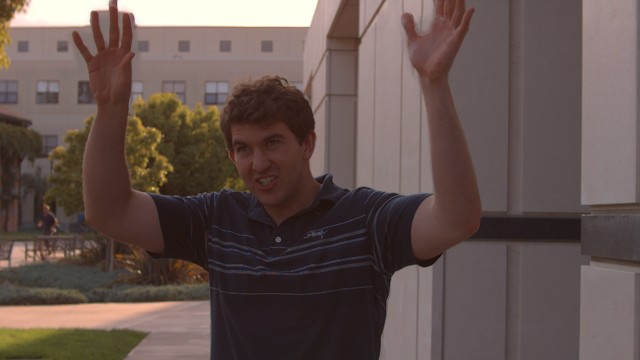}}\,\!
    \subfloat [MAP-HDR] {\includegraphics[width=3.5cm]{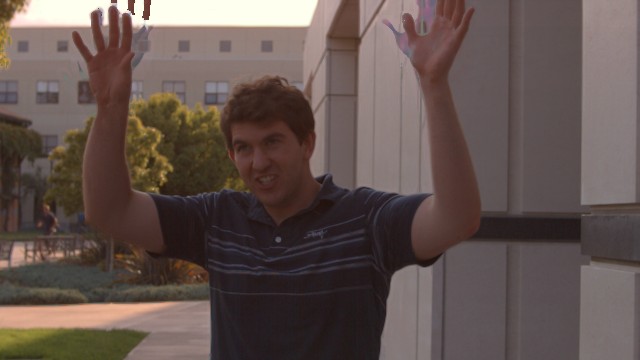}}\\

    \caption
    {
        Synthesized results of the 106th frame on the \emph{ThrowingTowel} sequence~\cite{kalantari_patch-based_2013} (top row) and the 16th frame of the \emph{WavingHands} sequence~\cite{kalantari_patch-based_2013} (bottom row).
    }
    \label{fig:breakdown}
\end{figure*}

While our method can do well in a wide variety of scenarios,
    there is a breaking point in terms of the amount of motion permissible
    between frames. Fig.~\ref{fig:breakdown} shows two such examples for the
    \emph{ThrowingTowel} and
    \emph{WavingHands} sequences~\cite{kalantari_patch-based_2013}, respectively. For both
    sequences, Kang {\it et al.}'s algorithm and
    MAP-HDR in Figs.~\ref{fig:breakdown}(b) and (e), respectively, produce ghosting artifacts. Because of large
    displacements, even after pyramidal decomposition, some corresponding pixels
    cannot be brought into a common local block, and thus neither optical flow
    nor kernel regression can find correct correspondences. 
    Mangiat {\it et al.}'s algorithm in Fig.~\ref{fig:breakdown}(c) provides significant amounts of artifacts, {\it e.g.}, on the moving arm and the towel in \emph{ThrowingTowel} and the waving
    hands in \emph{WavingHands}, due to the limitation of optical flow in handling large motions. On the other hand, Kalantari {\it et al}.'s algorithm in Fig.~\ref{fig:breakdown}(d) produces the least amount of artifacts. This is because their algorithm employs
    the PatchMatch algorithm~\cite{barnes_tog_2009} that is capable of handling
    huge displacements. By far, Kalantari {\it et al.}'s algorithm is the only
    one capable of generating high-quality videos in presence of huge motions,
    while our algorithm performs the best for reasonably large motions. However, notice that the proposed MAP-HDR requires substantially lower computational complexity than Kalantari {\it et al.}'s algorithm as will be discussed in Section~\ref{ssec:complexity}.

\subsection{Objective Video Quality Assessment}
\label{ssec:drivqm}

In addition to the subjective evaluation, we compare the MAP-HDR algorithm with
the state-of-the-art algorithms using five objective quality metrics: dynamic
range independent video quality metric (DRIVQM)~\cite{aydin_tog_2010}, logPSNR,
perceptually uniform extension to PSNR (puPSNR)~\cite{SPIE_Aydin_2008}, high
dynamic range-visible difference predictor (HDR-VDP)~\cite{TOG_Mantiuk_2011,
    JEI_Narwaria_2015}, and high dynamic range-visual quality metric
(HDR-VQM)~\cite{SPIC_Narwaria_2015}. DRIVQM for each frame estimates the
probability that the differences between two frames are to be noticed by a
careful viewer in each local region. The logPSNR and puPSNR metrics are
extensions of the peak signal-to-noise ratio (PSNR) by taking into account the
nonlinear perception of the human visual system to real-world luminance.
HDR-VDP estimates the probability at which an average human observer will
detect differences between the reference and the query images. In this work, we
use the $Q$ correlate of its version 2.2.1~\cite{JEI_Narwaria_2015}.    HDR-VQM
predicts the reconstruction quality of the HDR videos using perceptually
uniform encoding, sub-band decomposition, short- and long-term spatiotemporal
pooling, and the color information of the frames. Note that only DRIVQM and
HDR-VQM are dedicated to HDR video quality assessment, whereas the others are
developed for still images. Also, because these metrics require the original HDR
data as reference, we evaluate the performance only on \emph{Bridge2},
\emph{Hallway2}, \emph{ParkingLot}, and \emph{Students}.

\begin{figure}
    \centering

    \rotatebox[origin=c]{90}{\scriptsize \emph{Hallway2}}
    \subfloat {\includegraphics[width=1.9cm,valign=c]{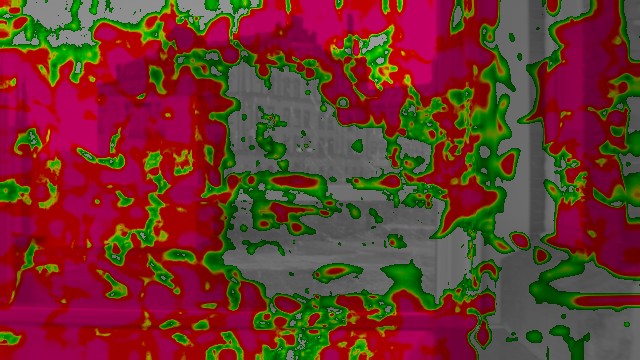}}\,\!
    \subfloat {\includegraphics[width=1.9cm,valign=c]{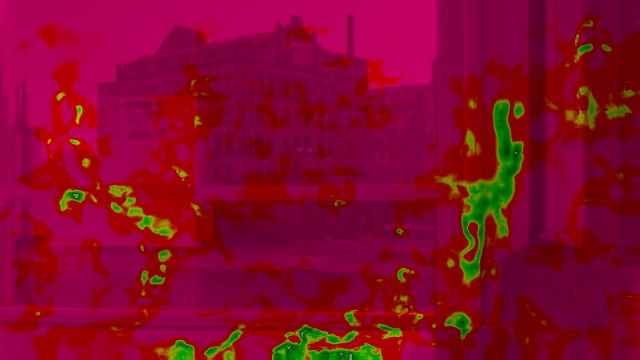}}\,\!
    \subfloat {\includegraphics[width=1.9cm,valign=c]{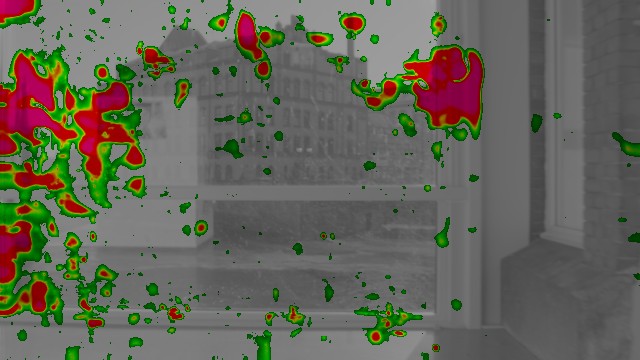}}\,\!
    \subfloat {\includegraphics[width=1.9cm,valign=c]{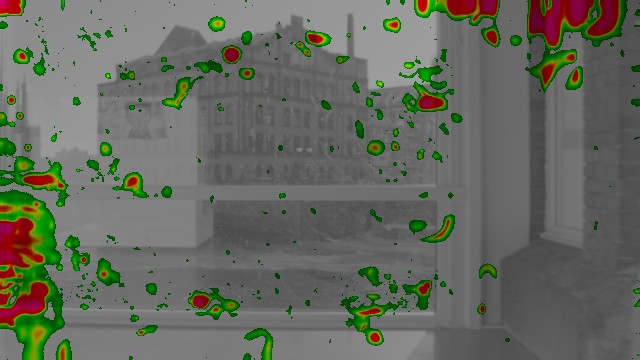}}\\
    \vspace{-2.5mm}

    \rotatebox[origin=c]{90}{\scriptsize \emph{Bridge2}}
    \subfloat {\includegraphics[width=1.9cm,valign=c]{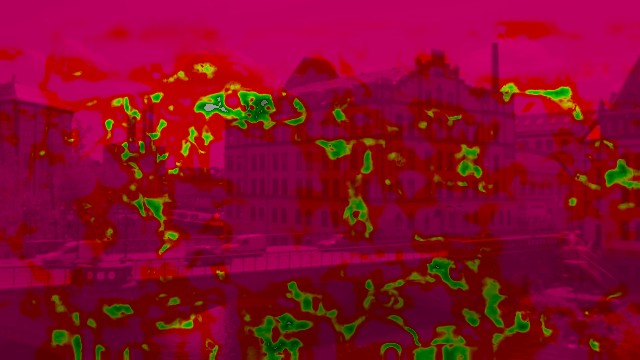}}\,\!
    \subfloat {\includegraphics[width=1.9cm,valign=c]{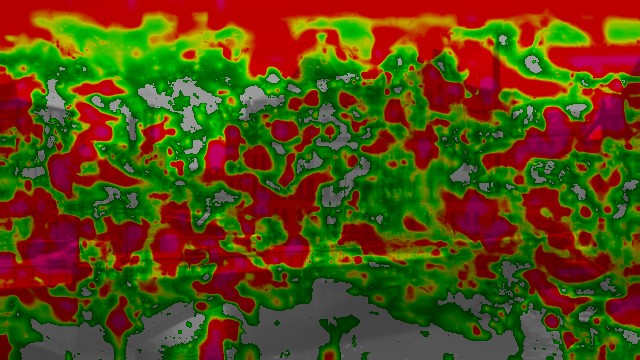}}\,\!
    \subfloat {\includegraphics[width=1.9cm,valign=c]{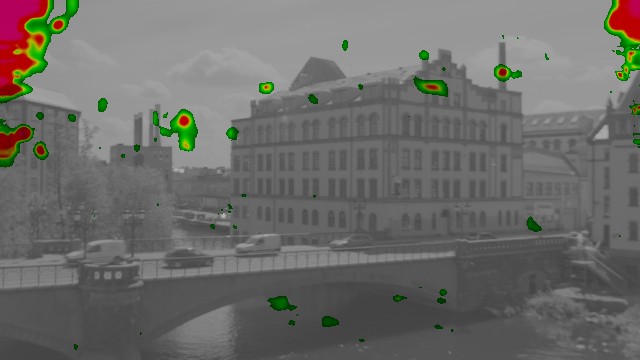}}\,\!
    \subfloat {\includegraphics[width=1.9cm,valign=c]{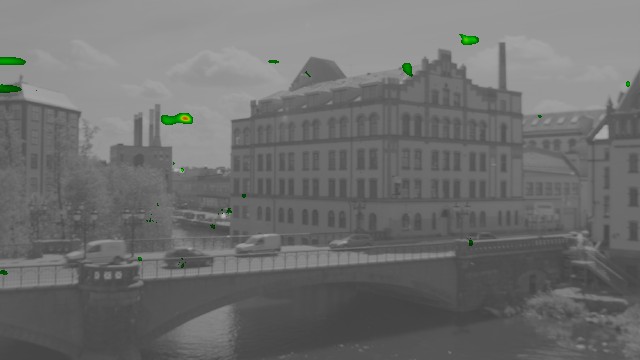}}\\
    \vspace{-2.5mm}

    \rotatebox[origin=c]{90}{\scriptsize \emph{ParkingLot}}
    \subfloat {\includegraphics[width=1.9cm,valign=c]{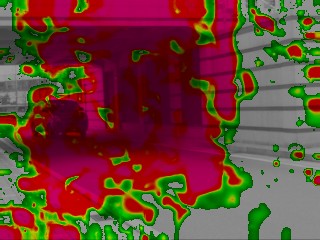}}\,\!
    \subfloat {\includegraphics[width=1.9cm,valign=c]{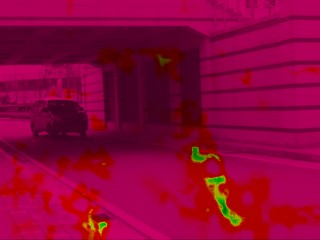}}\,\!
    \subfloat {\includegraphics[width=1.9cm,valign=c]{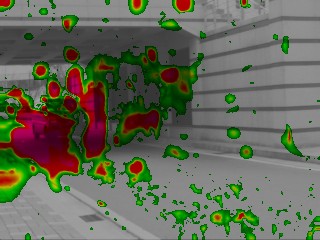}}\,\!
    \subfloat {\includegraphics[width=1.9cm,valign=c]{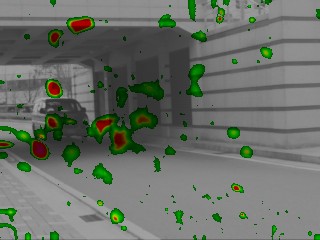}}\\
    \vspace{-2.5mm}

    \setcounter{subfigure}{0}
    \rotatebox[origin=c]{90}{\scriptsize \emph{Students}}\hspace{0.5mm}
    \subfloat [Kang {\it et al.}~\cite{kang_high_2003}] {\includegraphics[width=1.9cm,valign=c]{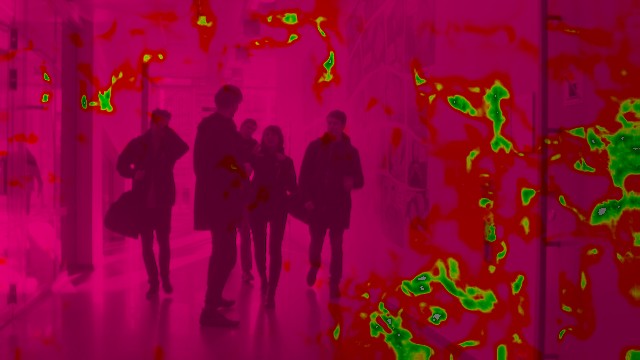}}\,\!
    \subfloat[Mangiat and Gibson~\cite{mangiat_spatially_2011}]  {\includegraphics[width=1.9cm,valign=c]{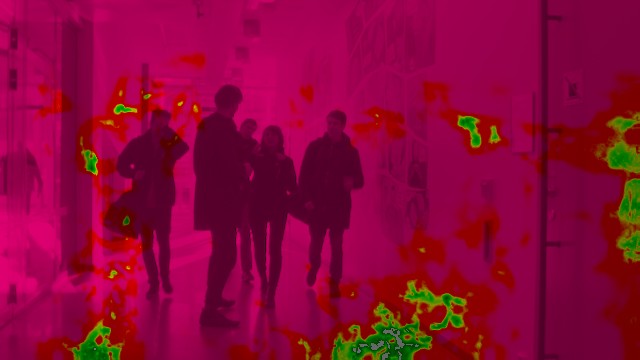}}\,\!
    \subfloat [Kalantari {\it et al.}~\cite{kalantari_patch-based_2013}] {\includegraphics[width=1.9cm,valign=c]{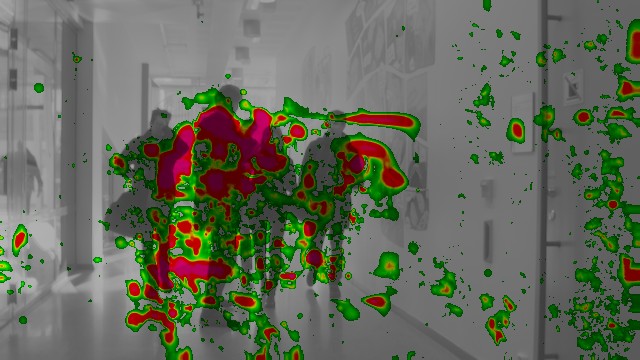}}\,\!
    \subfloat [MAP-HDR] {\includegraphics[width=1.9cm,valign=c]{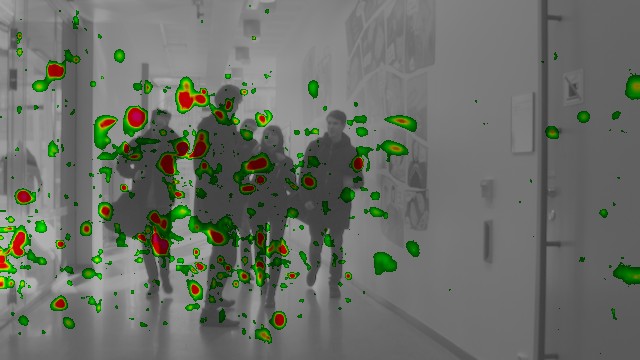}}\\

    \subfloat[] {\includegraphics[scale=0.85]{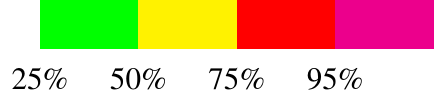}}

    \caption
    {
        DRIVQM assessment on the 140th frame of \emph{Hallway2} (top row), the 10th frame of \emph{Bridge2} (second row), the 65th frame of \emph{ParkingLot} (third row), and the 21st frame of \emph{Students} (bottom row). The colormap depicts the predicted visible differences, {\it i.e.}, the estimated  probability shown in (e), that an observer would notice the differences between the synthesized frames and the references.
    }
    \label{fig:drivqm}
\end{figure}

Fig.~\ref{fig:drivqm} compares the selected frames of the DRIVQM assessment
results on the test sequences. Kang {\it et al.}'s algorithm in
Fig.~\ref{fig:drivqm}(a) provides severe visible differences, especially on the
\emph{Bridge2} and \emph{Students} sequences, since it composes parts of two
different frames together. Mangiat and Gibson's algorithm in
Fig.~\ref{fig:drivqm}(b) produces visible differences throughout the frames on
the \emph{Hallway2}, \emph{ParkingLot}, and \emph{Studnets} sequences, since
its HDR filtering severely blurs the whole frame. Kalantari {\it et al.}'s
algorithm in Fig.~\ref{fig:drivqm}(c) produces much lower visible differences
on the \emph{Hallway2} and \emph{Brigde2} sequences but poor results on the
\emph{ParkingLot} and \emph{Students} sequences. On the contrary, we can see
that the MAP-HDR algorithm in Fig.~\ref{fig:drivqm}(d) produces high-quality
video results with significantly fewer visible differences than all
state-of-the-art algorithms.

\begin{table*}
    \renewcommand{\arraystretch}{1.1}
    \setlength{\tabcolsep}{0.9mm}
    \caption
    {
        Objective assessment of the HDR video synthesis performance using four metrics: logPSNR, puPSNR~\cite{SPIE_Aydin_2008}, HDR-VDP~\cite{TOG_Mantiuk_2011, JEI_Narwaria_2015}, and HDR-VQM~\cite{SPIC_Narwaria_2015}. Bold-faced numbers denote the highest scores for each metric on each sequence. For all metrics, higher values indicate better quality.
    }
    \centering
    \scalebox{0.8}{%
        \begin{tabular} {c||c|c|c|c||c|c|c|c||c|c|c|c||c|c|c|c}
            \hline
            \hline
            & \multicolumn{4}{c||}{logPSNR} & \multicolumn{4}{c||}{puPSNR} & \multicolumn{4}{c||}{HDR-VDP} & \multicolumn{4}{c}{HDR-VQM} \\
            \cline{2-17}
            & \emph{Bridge2} & \emph{Hallway2} & \emph{ParkingLot} & \emph{Students} & \emph{Bridge2} & \emph{Hallway2} & \emph{ParkingLot} & \emph{Students} & \emph{Bridge2} & \emph{Hallway2} & \emph{ParkingLot} & \emph{Students} & \emph{Bridge2} & \emph{Hallway2} & \emph{ParkingLot} & \emph{Students} \\
            \hline
            Kang {\it et al.}       & 25.56 & 34.06 & 12.78 & 17.23    & 40.66 & 33.88 & 23.57  & 18.81    & 59.26 & 59.22 & 51.02 & 37.58    & 0.7435 & 0.7417 & 0.7829 & 0.5852 \\
            \hline
            Mangiat and Gibson      & 31.85 & 35.06 & 29.89 & {\bf 22.66}    & 44.53 & 35.18 & 34.19  & {\bf 26.54}    & {\bf 61.56} & 57.91 & 55.78 & 50.07    & 0.8906 & 0.6913 & 0.3880 & 0.5665 \\
            \hline
            Kalantari {\it et al.}  & 36.01 & 34.98 & 29.30 & 21.20    & 45.56 & 35.26 & {\bf 38.11} & 25.47    & 60.09 & 62.91 & 65.92 & {\bf 51.21}    & 0.9446 & 0.8840 & 0.7904 & {\bf 0.6677} \\
            \hline
            MAP-HDR                 & {\bf 36.48} & {\bf 35.81} & {\bf 31.05} & 20.69    & {\bf 48.05} & {\bf 36.08} & 36.93 & 24.96    & 59.33 & {\bf 63.91} & {\bf 66.07} & 51.06    & {\bf 0.9863} & {\bf 0.9242} & {\bf 0.8418} & 0.6518 \\
            \hline
            \hline
        \end{tabular}}
    \label{table:objective_assessment}
\end{table*}

Table~\ref{table:objective_assessment} quantitatively compares the
reconstruction performance of the MAP-HDR algorithm with those of the
state-of-the-art algorithms over four test sequences.
For each metric on each sequence, the highest scores that indicate the best
results are bold-faced. First, logPSNR and puPSNR measure the quality of pixel
value reconstruction. Since the proposed MAP-HDR effectively captures motion
information via the multiscale adaptive kernel regression algorithm, it
provides the highest or close to highest scores on all the test videos in terms
of logPSNR and puPSNR, except that it provides the third highest scores on
\emph{Students}. Second, HDR-VDP and HDR-VQM estimate perceptual differences
between two videos. On \emph{Bridge2}, \emph{Hallway2}, and \emph{ParkingLot},
the proposed MAP-HDR provides the best performance in terms of both HDR-VDP and
HDR-VQM, except for \emph{Bridge2} where it provides the third highest HDR-VDP
score. We notice that the~\emph{Students} video contains a local regions with
extremely high dynamic range of irradiance values, and fitting data with large
variations is an inherent limitation of polynomial regression. Therefore, the
MAP-HDR provides lower scores on the \emph{Students} video, although it
provides the best visual quality as shown in Fig.~\ref{fig:students}.
Nevertheless, MAP-HDR achieves comparable scores with other algorithms. To
summarize, capturing both local structure and temporal motion by solving
optimization problems effectively, the proposed MAP-HDR provides overall
high-quality videos in these quality metrics.

\subsection{Computational Complexity Analysis}
\label{ssec:complexity}

\begin{table*}
    \renewcommand{\arraystretch}{1.1}
    \setlength{\tabcolsep}{1.9mm}
    \caption
    {
        Comparison of the computational complexities. $N$ and $N_f$ denote the
        number of pixels in a frame and foreground parts, respectively, $P$ is
        the block sizes (different in each algorithm), $M$ denotes the
        number of parameters in the kernel regression, $J_{\mathrm{SOR}}$ and
        $J_{\mathrm{BFGS}}$ are the iteration numbers in the SOR and BFGS
        procedures, respectively, and $J_{\mathrm{MAX}} =
        \max\{J_{\mathrm{SOR}}, J_\mathrm{O}J_\mathrm{MC}\}$, where
        $J_\mathrm{O}$ and $J_\mathrm{MC}$ denote the number of outer loops and
        matrix completion iterations in MAP-HDR, respectively.
        $J_{\mathrm{REC}}$ denotes the number of outer iterations in
        Kalantari's algorithm. We also include typical parameter settings
        and a numerical comparison against a $640\times
        480$ gray image, giving $N=307200$, and we let
        $N_f=0.05N=15360$.
    }
    \centering
    \begin{tabular} {c||c|c|c|c}
        \hline
        \hline
        &  Kang {\it et al.}  &  Mangiat and Gibson  &  Kalantari {\it et al.}  &  MAP-HDR\\
        \hline
        \hline
        Computational Complexity &  $\mathcal{O}(P^2N)$  &  $\cO(P^2N)$  & $\mathcal{O}(J_{\mathrm{REC}}P^2N\log N + J_{\mathrm{SOR}}N)$  & $\mathcal{O}(N_fJ_{\mathrm{BFGS}}M^2P^2+J_{\mathrm{MAX}}N)$ \\
        \hline
        Typical parameter settings & $P=5$ & $P=16$ & $P=7,J_{\mathrm{REC}}=15$ & $P=7,J_{\mathrm{BFGS}}=10,M=10,J_{\mathrm{MAX}}=500$\\
        \hline
        Resulting asymptotic costs & $7.68\times 10^6$ & $7.86\times 10^7$ & $4.12\times 10^9$ & $7.63\times 10^8$\\
        \hline
        \hline
    \end{tabular}
    \label{table:complexity}
\end{table*}

We analyze the computational complexity of synthesizing HDR videos with an
$N$-pixel frame. For Kang {\it et al.}'s algorithm~\cite{kang_high_2003},
gradient calculation and image warping via linear interpolation are linear in
complexity, which is much lower than other parts of the algorithm, and we
hereafter omit their contributions. The critical part of Kang {\it et al.}'s
algorithm is hierarchical computation of Lucas-Kanade optical
flow~\cite{lucas_ijcai_1981}, which has a complexity of $\mathcal{O}(P^2N)$,
where $P\times P$ is the window size for computing the local structure tensor
matrix. Thus, the complexity of Kang {\it et al.}'s algorithm is
$\mathcal{O}(P^2N)$.

Mangiat and Gibson's algorithm~\cite{mangiat_spatially_2011} involves two
critical parts: block-based motion estimation and HDR filtering. The former has
a complexity of $\mathcal{O}(P^2N)$~\cite{nie_tip_2002}, where $P\times P$ is
the block size for block matching. The latter has the same complexity as
bilateral filtering, {\it i.e.}, $\mathcal{O}(P^2N)$. Thus, Mangiat and
Gibson's algorithm has an overall complexity of $\mathcal{O}(P^2N)$.

We compute the complexity of Kalantari {\it et al.}'s
algorithm~\cite{kalantari_patch-based_2013} by dividing it into three steps as
follows.
\begin{enumerate}
    \item Initial motion estimation: The critical part of this step is optical
    flow estimation~\cite{liu2009beyond}, which mainly involves successive over-relaxation (SOR) iterations
    for solving linear systems. Since each iteration is linear in $N$, the
    complexity of this step is $\mathcal{O}(J_{\mathrm{SOR}}N)$, where
    $J_{\mathrm{SOR}}$ denotes the number of SOR iterations.

    \item Search window map computation: This step consists of running
    PatchMatch~\cite{barnes_tog_2009} iteratively for the correspondence estimation. Since PatchMatch has a complexity of
    $\mathcal{O}(P^2N\log N)$~\cite{barnes_tog_2009} for $P\times P$ patches,
    the complexity of this step is also $\mathcal{O}(P^2N\log N)$.

    \item HDR video reconstruction: This is an iterative procedure, and each
    iteration consists of two steps, {\it i.e.}, SearchVote and AlphaBlend. In the SearchVote procedure, PatchMatch is executed to find correspondences between
    reference frame and neighboring frames, which is of $\mathcal{O}(P^2N\log
    N)$ complexity, while AlphaBlend merges aligned frames together, which is of $\mathcal{O}(N)$ complexity. Overall, the
    complexity is $\mathcal{O}(J_{\mathrm{REC}}(P^2N\log N +
    N))=\mathcal{O}(J_{\mathrm{REC}}P^2N\log N)$, where $J_{\mathrm{REC}}$ is
    the number of outer iterations.
\end{enumerate}
Therefore, the complexity of Kalantari {\it et al.}'s algorithm is
$\mathcal{O}(J_{\mathrm{REC}}P^2N\log N + J_{\mathrm{SOR}}N)$.

The MAP-HDR algorithm consists of two stages: matrix completion for computing
the background parts and kernel regression for computing the foreground parts.
The first stage turns out to be of $\mathcal{O}(J_{\mathrm{O}}J_{MC}N)$
complexity, where $J_{MC}$ and $J_{\mathrm{O}}$ denote the number of iterations
in the matrix completion step~\cite{wen_solving_2012} and the number of outer
loops, respectively.  Thanks to the low-rank property of the background matrix
(the rank is at most $3$), the QR decomposition procedure becomes trivial as
noted in~\cite{wen_solving_2012}, and hence the complexity of each iteration is
linear in $N$. The critical step of the second stage is the BFGS optimization
in~(\ref{eqn:objective}), whose critical step is in turn the computation of
$\bJ$ in~(\ref{eqn:gradient}), which is of complexity $\mathcal{O}(M^2P^2)$, where
$M$ and $P$ are the number of parameters for kernel regression and the width of
local sample block, respectively.  Therefore, the second stage is of complexity
$\mathcal{O}(N_fJ_{\mathrm{BFGS}}M^2P^2)$, where $N_f$ and $J_{\mathrm{BFGS}}$
denote the number of foreground pixels and the number of iterations in the BFGS
optimization step, respectively. In addition, the optical flow
estimation~\cite{liu2009beyond} in the pyramidal implementation is of
complexity $\mathcal{O}(J_{\mathrm{SOR}}N)$ as in Kalantari {\it et al.}'s
algorithm. Overall, the complexity of the MAP-HDR algorithm is
$\mathcal{O}(N_fJ_{\mathrm{BFGS}}M^2P^2+J_{\mathrm{MAX}}N)$, where
$J_{\mathrm{MAX}} = \max\{J_{\mathrm{SOR}}, J_{\mathrm{O}}J_{\mathrm{MC}}\}$.

We summarize the complete complexities in Table~\ref{table:complexity}.  We see
that both Kang {\it et al.}'s and Mangiat and Gibson's methods have relatively
lower complexities; however, both methods are prone to introducing serious
artifacts when mis-registration happens as shown in
Figs~\ref{fig:fire}--\ref{fig:parkinglot}. Let us compare complexities of
Kalantari {\it et al.}'s algorithm and the MAP-HDR algorithm quantitatively. For Kalantari
{\it et al.}'s algorithm with $P=7$, summing up $J_{\mathrm{REC}}$'s in each
scale gives an equivalent $J_{\mathrm{REC}}\approx 15$, and usually
$J_{\mathrm{SOR}}N\ll J_{\mathrm{REC}}P^2N\log N$. In contrast, in the MAP-HDR
algorithm, typically $J_{\mathrm{BFGS}}\leq 10$, $M=10$, $P=7$, $N_f\ll N$, and
$J_{\mathrm{MAX}}$ typically varies from several hundreds to a thousand. 
 Because
    our algorithm produces better results than competing methods in most (but not all) cases, it achieves a better complexity-performance trade-off.

%% file: section6.tex
\section{Conclusion}
\label{sec:conclusion}
We develop a MAP estimation-based approach to HDR video synthesis from a series
of LDR frames taken with alternating exposure times. By choosing physically
meaningful priors, {\it e.g.}, low-rank background and the Ising model, the MAP estimation can be reduced to a tractable optimization problem, the solution to which provides aligned background regions
as well as foreground irradiance that implicitly captures motion information. To capture motion
information of the foreground regions, we further develop a multiscale adaptive
kernel regression algorithm. Finally, HDR frame synthesis simply follows by
combining the foreground with the background. Experimental results on
challenging video sequences demonstrate the effectiveness of our algorithm in
obtaining high-quality HDR video sequences. Some of the important directions for future work are to incorporate more accurate noise models, such as those mentioned in~\cite{hasinoff_noise-optimal_2010,granados_optimal_2010}, and to extend the framework towards handling very large object and camera motion as in \cite{kalantari_patch-based_2013}.

%% file: appendix1.tex
\section{Derivation of the Estimation in (\ref{eqn:final})}
\label{sec:appendix_map}
Since $\bS$ is the support of $\bF$, by the Bayes formula, the posterior probability $f(\bS, \bB, \bF|\bD) = \frac{f(\bD, \bB, \bF|\bS)f(\bS)}{f(\bD)}$ in (\ref{eqn:map}) is given by
\begin{align*}
    f(\bS, \bB, \bF|\bD) & = f(\bD, \bB, \cP_\bS(\bF)|\bS)f(\bS)  \\
    & = f(\cP_{\bS^\mathsf{c}}(\bD), \cP_\bS(\bD), \bB, \cP_\bS(\bF)|\bS)f(\bS),
\end{align*}
where we use the identity $\bD = \cP_{\bS^\mathsf{c}}(\bD) + \cP_\bS(\bD)$. Then, based on the assumption that the foreground $\bF$ and background $\bB$ are independent, from (\ref{eqn:asum1}) and (\ref{eqn:asum2}), we have
\begin{align}
    f(\bS, \bB, \bF|\bD) &= f(\cP_{\bS^\mathsf{c}}(\bD), \bB|\bS) f(\cP_\bS(\bF), \cP_{\bS}(\bD)|\bS)f(\bS) \nonumber\\
    &= f(\cP_{\bS^\mathsf{c}}(\bD), \bB) f(\cP_\bS(\bF), \cP_{\bS}(\bD))f(\bS).
    \label{eqn:ap_asum3}
\end{align}
Taking the logarithm of both sides of (\ref{eqn:ap_asum3}) yields
\begin{align*}
    \log f(\bS, \bB, \bF|\bD) &= \log f(\cP_{\bS^\mathsf{c}}(\bD), \bB) \\
    &{}\quad + \log f(\cP_\bS(\bF), \cP_{\bS}(\bD)) + \log f(\bS) \\
    &= \log f(\cP_{\bS^\mathsf{c}}(\bD)|\bB) + \log f(\bB) \\
    &{}\quad + \log f(\cP_\bS(\bF), \cP_\bS(\bD)) + \log f(\bS).
\end{align*}
Finally, since
$$
\underset{\bS,\bB,\bF}{\arg\max}\; f(\bS, \bB, \bF|\bD) = \underset{\bS,\bB,\bF}{\arg\max}\; \log f(\bS, \bB, \bF|\bD),
$$
we obtain (\ref{eqn:final}).

%% file: appendix2.tex
\section{Derivation of the Gradient $\nabla\mathcal{C}$ in (\ref{eqn:gradient})}
\label{sec:appendix_gradient}

To illustrate, we consider each term in (\ref{eqn:objective})
separately as
\begin{align*}
    \mathcal{C}_1 &:= (\by-\bX\widehat{\bm{\beta}})^T\bm{\Lambda}(\by - \bX\widehat{\bm{\beta}})+(\by-t\bone)^T\widetilde{\bm{\Lambda}}(\by-t\bone), \\
    \mathcal{C}_2 &:= \|\widehat{\bm{\beta}}\|_2^2,\\
    \mathcal{C}_3 &:= \|\bR\|_F^2.
\end{align*}

As in (\ref{eqn:objective}), let $\bT = (\bX^T\bLambda\bX+\varepsilon\bI)^{-1}\bX^T$ and $\bJ = \bX\bT$. Then, since $\mathrm{d}{\bA^{-1}} = -\bA^{-1}(\mathrm{d}\bA)\bA^{-1}$~\cite{magnus1995matrix}, we have
\begin{align*}
    \mathrm{d}\bT &= -(\bX^T\bm{\Lambda}\bX + \varepsilon\bI)^{-1}\bX^T(\mathrm{d}\bm{\Lambda})\bX(\bX^T\bm{\Lambda}\bX + \varepsilon\bI)^{-1}\bX^T\\
    &= -\bT(\mathrm{d}\bm{\Lambda})\bJ.
\end{align*}
Therefore,
\begin{align*}
    \mathrm{d}\widehat{\bm{\beta}} &= -\bT(\mathrm{d}\bm{\Lambda})\bJ\bm{\Lambda}\by + \bT(\mathrm{d}\bm{\Lambda})\by\\
    &= \bT(\mathrm{d}\bm{\Lambda})(\by - \bX\widehat{\bm{\beta}})
\end{align*}
and $\mathrm{d}\widehat{y}_i = \be_i^T\mathrm{d}(\bX\widehat{\bm{\beta}}) = \sum_{j\in\mathcal{G}}J_{ij}(y_j-\widehat{y}_j)\mathrm{d} k_j$.
Also, we have
\begin{align*}
    \mathrm{d} J_{ii} &= \be_i^T(\mathrm{d}\bJ)\be_i = -\be_i^T\bJ(\mathrm{d}\bm{\Lambda})\bJ\be_i= -\sum_{j\in\mathcal{G}}J_{ij}^2\mathrm{d} k_j.
\end{align*}
Next, by the chain rule~\cite{magnus1995matrix}, we can obtain
\begin{align*}
    \mathrm{d} k_i &= k_i\mathrm{d} \!\left(-\frac{1}{2}(\bx_i-\bx)^T\bR^T\bR(\bx_i-\bx)\right) \\
    &= -k_i \left\langle \bR(\bx_i-\bx)(\bx_i-\bx)^T, \mathrm{d}\bR \right\rangle\!,
\end{align*}
where $\langle\bA,
\bB\rangle:=\mathrm{tr}(\bA^T\bB)$ denotes the Frobenius inner product of two matrices $\bA$ and $\bB$. Similarly, for $\tilde{k}_i$,
\begin{align*}
    \mathrm{d}\tilde{k}_i&= -\tilde{k}_i \left\langle \bR(\bx_i-\bx)(\bx_i-\bx)^T, \mathrm{d}\bR \right\rangle\!.
\end{align*}
Thus,
\begin{align*}
    \mathrm{d}\mathcal{C}_1 &= \mathrm{d}\!\left\{\sum_{i}k_i(y_i-\widehat{y}_i)^2+\tilde{k}_i(y_i-t)^2\right\}\\
    &= \sum_{i}(y_i-\widehat{y}_i)^2\mathrm{d}k_i + k_i\mathrm{d}(y_i-\widehat{y}_i)^2+(y_i-t)^2\mathrm{d}\tilde{k}_i\\
    &= \left\langle\!\bR\Bigg\{2\sum_{i,j}k_ik_jJ_{ij}(y_i-\widehat{y}_i)(y_j-\widehat{y}_j)(\bx_j-\bx)(\bx_j-\bx)^T\right. \\
    &{}\quad \left.- \sum_{i}\left(k_iz_i^2+\tilde{k}_i\tilde{z}_i^2\right)(\bx_i-\bx)(\bx_i-\bx)^T\Bigg\}\!, \mathrm{d}\bR\right\rangle\!.
\end{align*}

Also, we can derive $\mathrm{d}\mathcal{C}_2$ and $\mathrm{d}\mathcal{C}_3$ as
\begin{align*}
    \mathrm{d}\mathcal{C}_2 &= \left\langle -2\bR\sum_{i,j}\widehat{\beta}_ik_jT_{ij}(y_j-\widehat{y}_j)(\bx_j-\bx)(\bx_j-\bx)^T, \mathrm{d}\bR \right\rangle
\end{align*}
and $\mathrm{d}\mathcal{C}_3 = \left\langle 2\bR, \mathrm{d}\bR \right\rangle$, respectively.

Since $\mathrm{d}\mathcal{C} = \mathrm{d}\mathcal{C}_1 + \lambda\mathrm{d}\mathcal{C}_2 + \mu\mathrm{d}\mathcal{C}_3$, and since $\bR$ is upper triangular, so is $\mathrm{d}\bR$; and $\nabla\mathcal{C}$ is thus given by (\ref{eqn:gradient}).

%% file: main.bbl
\begin{thebibliography}{10}
\providecommand{\url}[1]{#1}
\csname url@samestyle\endcsname
\providecommand{\newblock}{\relax}
\providecommand{\bibinfo}[2]{#2}
\providecommand{\BIBentrySTDinterwordspacing}{\spaceskip=0pt\relax}
\providecommand{\BIBentryALTinterwordstretchfactor}{4}
\providecommand{\BIBentryALTinterwordspacing}{\spaceskip=\fontdimen2\font plus
\BIBentryALTinterwordstretchfactor\fontdimen3\font minus
  \fontdimen4\font\relax}
\providecommand{\BIBforeignlanguage}[2]{{%
\expandafter\ifx\csname l@#1\endcsname\relax
\typeout{** WARNING: IEEEtran.bst: No hyphenation pattern has been}%
\typeout{** loaded for the language `#1'. Using the pattern for}%
\typeout{** the default language instead.}%
\else
\language=\csname l@#1\endcsname
\fi
#2}}
\providecommand{\BIBdecl}{\relax}
\BIBdecl

\bibitem{HDRI_BOOK_Reinhard}
E.~Reinhard, G.~Ward, S.~Pattanaik, P.~Debevec, W.~Heidrich, and K.~Myszkowski,
  \emph{High Dynamic Range Imaging: Acquisition, Display, and Image-Based
  Lighting}, 2nd~ed.\hskip 1em plus 0.5em minus 0.4em\relax San Mateo, CA:
  Morgan Kaufmann Publishers, 2010.

\bibitem{nayar_adaptive_2003}
S.~K. Nayar and V.~Branzoi, ``Adaptive dynamic range imaging: Optical control
  of pixel exposures over space and time,'' in \emph{Proc. {IEEE} Int. Conf.
  Comput. Vis.}, Oct. 2003, pp. 1168--1175.

\bibitem{tocci_2011_TOG}
M.~D. Tocci, C.~Kiser, N.~Tocci, and P.~Sen, ``A versatile {HDR} video
  production system,'' \emph{ACM Trans. Graphics}, vol.~30, no.~4, pp.
  41:1--10, Jul. 2011.

\bibitem{kronander2014unified}
J.~Kronander, S.~Gustavson, G.~Bonnet, A.~Ynnerman, and J.~Unger, ``A unified
  framework for multi-sensor {HDR} video reconstruction,'' \emph{Signal
  Process.: Image Commun.}, vol.~29, no.~2, pp. 203--215, Feb. 2014.

\bibitem{debevec_recovering_2008}
P.~E. Debevec and J.~Malik, ``Recovering high dynamic range radiance maps from
  photographs,'' in \emph{Proc. {ACM} SIGGRAPH}, Aug. 1997, pp. 369--378.

\bibitem{sen_robust_2012}
P.~Sen, N.~K. Kalantari, M.~Yaesoubi, S.~Darabi, D.~B. Goldman, and
  E.~Shechtman, ``Robust patch-based {HDR} reconstruction of dynamic scenes.''
  \emph{{ACM} Trans. Graphics}, vol.~31, no.~6, pp. 203:1--11, Nov. 2012.

\bibitem{hu_hdr_2013}
J.~Hu, O.~Gallo, K.~Pulli, and X.~Sun, ``{HDR} deghosting: How to deal with
  saturation?'' in \emph{Proc. {IEEE} Conf. Comput. Vis. Pattern Recognit.},
  Jun. 2013, pp. 1163--1170.

\bibitem{lee_ghost-free_2014}
C.~Lee, Y.~Li, and V.~Monga, ``Ghost-free high dynamic range imaging via rank
  minimization,'' \emph{{IEEE} Signal Process. Lett.}, vol.~21, no.~9, pp.
  1045--1049, Sep. 2014.

\bibitem{oh_robust_2014}
T.-H. Oh, J.-Y. Lee, Y.-W. Tai, and I.~S. Kweon, ``Robust high dynamic range
  imaging by rank minimization,'' \emph{{IEEE} Trans. Pattern Anal. Mach.
  Intell.}, vol.~37, no.~6, pp. 1219--1232, Jun. 2015.

\bibitem{CGF_Tursun_2015}
O.~T. Tursun, A.~O. Aky\"{u}z, A.~Erdem, and E.~Erdem, ``The state of the art
  in {HDR} deghosting: A survey and evaluation,'' \emph{Comput. Graph. Forum},
  vol.~34, no.~2, pp. 683--707, May 2015.

\bibitem{kang_high_2003}
S.~B. Kang, M.~Uyttendaele, S.~Winder, and R.~Szeliski, ``High dynamic range
  video,'' \emph{{ACM} Trans. Graphics}, vol.~22, no.~3, pp. 319--325, Jul.
  2003.

\bibitem{lucas_ijcai_1981}
B.~D. Lucas and T.~Kanade, ``An iterative image registration technique with an
  application to stereo vision,'' in \emph{Proc. 7th Int. Joint Conf.
  Artificial Intell.}, 1981, pp. 674--679.

\bibitem{liu2009beyond}
C.~Liu, ``Beyond pixels: Exploring new representations and applications for
  motion analysis,'' Ph.D. dissertation, Massachusetts Institute of Technology,
  Cambridge, MA, May 2009.

\bibitem{unger2007high}
J.~Unger and S.~Gustavson, ``High-dynamic-range video for photometric
  measurement of illumination,'' in \emph{Proc. SPIE Sensors, Cameras, and
  Systems for Scientific/Industrial Applications VIII}, vol. 6501, Feb. 2007,
  pp. 6501:1--10.

\bibitem{mangiat_high_2010}
S.~Mangiat and J.~Gibson, ``High dynamic range video with ghost removal,'' in
  \emph{Proc. SPIE Applications of Digital Image Processing XXXIII}, vol. 7798,
  Aug. 2010, pp. 779\,812:1--8.

\bibitem{horn_mit_1980}
B.~K. Horn and B.~G. Schunck, ``Determining optical flow,'' \emph{Artificial
  Intell.}, vol.~17, pp. 185--203, Apr. 1981.

\bibitem{mangiat_spatially_2011}
S.~Mangiat and J.~Gibson, ``Spatially adaptive filtering for registration
  artifact removal in {HDR} video,'' in \emph{IEEE Int. Conf. Image Process.},
  Sep. 2011, pp. 1317--1320.

\bibitem{mangiat_milcom_2011}
------, ``Inexpensive high dynamic range video for large scale security and
  surveillance,'' in \emph{Proc. Int. Conf. Military Commun.}, Nov. 2011, pp.
  1772--1777.

\bibitem{kalantari_patch-based_2013}
N.~K. Kalantari, E.~Shechtman, C.~Barnes, S.~Darabi, D.~B. Goldman, and P.~Sen,
  ``Patch-based high dynamic range video,'' \emph{{ACM} Trans. Graphics},
  vol.~32, no.~6, pp. 202:1--8, Nov. 2013.

\bibitem{ICIP_Li_2015}
Y.~Li, C.~Lee, and V.~Monga, ``A {MAP} estimation framework for {HDR} video
  synthesis,'' in \emph{Proc. IEEE Int. Conf. Image Process.}, Sep. 2015, pp.
  2219--2223.

\bibitem{takeda_kernel_2007}
H.~Takeda, S.~Farsiu, and P.~Milanfar, ``Kernel regression for image processing
  and reconstruction,'' \emph{{IEEE} Trans. Image Process.}, vol.~16, no.~2,
  pp. 349--366, Feb. 2007.

\bibitem{takeda_super-resolution_2009}
H.~Takeda, P.~Milanfar, M.~Protter, and M.~Elad, ``Super-resolution without
  explicit subpixel motion estimation,'' \emph{{IEEE} Trans. Image Process.},
  vol.~18, no.~9, pp. 1958--1975, Sep. 2009.

\bibitem{schaal1998constructive}
S.~Schaal and C.~G. Atkeson, ``Constructive incremental learning from only
  local information,'' \emph{Neural Computation}, vol.~10, no.~8, pp.
  2047--2084, Nov. 1998.

\bibitem{blake_markov_2011}
A.~Blake, P.~Kohli, and C.~Rother, \emph{Markov Random Fields for Vision and
  Image Processing}.\hskip 1em plus 0.5em minus 0.4em\relax Cambridge, MA: MIT
  Press, 2011.

\bibitem{cevher_sparse_2010}
V.~Cevher, P.~Indyk, L.~Carin, and R.~G. Baraniuk, ``Sparse signal recovery and
  acquisition with graphical models,'' \emph{{IEEE} Signal Process. Mag.},
  vol.~27, no.~6, pp. 92--103, Nov. 2010.

\bibitem{cevher_sparse_2008}
V.~Cevher, M.~F. Duarte, C.~Hegde, and R.~Baraniuk, ``Sparse signal recovery
  using {Markov} random fields,'' in \emph{Proc. Neural Inf. Process. Syst.},
  Dec. 2009, pp. 257--264.

\bibitem{TPAMI_Kolmogorov_2004}
V.~Kolmogorov and R.~Zabih, ``What energy functions can be minimized via graph
  cuts?'' \emph{{IEEE} Trans. Pattern Anal. Mach. Intell.}, vol.~26, no.~2, pp.
  147--159, Feb. 2004.

\bibitem{code_nlopt}
\BIBentryALTinterwordspacing
S.~G. Johnson. The {NLopt} nonliear optimization package. [Online]. Available:
  \url{http://ab-initio.mit.edu/nlopt}
\BIBentrySTDinterwordspacing

\bibitem{nocedal_mc_1980}
J.~Nocedal, ``Updating quasi-newton matrices with limited storage,''
  \emph{Math. Comput.}, vol.~35, pp. 773--782, 1980.

\bibitem{liu_mp_1989}
D.~C. Liu and J.~Nocedal, ``On the limited memory {BFGS} method for large scale
  optimization,'' \emph{Math. Program.}, vol.~45, no.~3, pp. 503--528, Dec.
  1989.

\bibitem{nocedal2006numerical}
J.~Nocedal and S.~J. Wright, \emph{Numerical Optimization}, 2nd~ed.\hskip 1em
  plus 0.5em minus 0.4em\relax New York, NY: Springer, 2006.

\bibitem{lee_rate-distortion_2012}
C.~Lee and C.-S. Kim, ``Rate-distortion optimized layered coding of high
  dynamic range videos,'' \emph{J. Vis. Commun. Image R.}, vol.~23, no.~6, pp.
  908--923, Aug. 2012.

\bibitem{brox_high_2004}
T.~Brox, A.~Bruhn, N.~Papenberg, and J.~Weickert, ``High accuracy optical flow
  estimation based on a theory for warping,'' in \emph{Proc. European Conf.
  Comput. Vis.}, May 2004, pp. 25--36.

\bibitem{bruhn2005lucas}
A.~Bruhn, J.~Weickert, and C.~Schn{\"o}rr, ``{Lucas/Kanade} meets
  {Horn/Schunck}: Combining local and global optic flow methods,'' \emph{Int.
  J. Comput. Vis.}, vol.~61, no.~3, pp. 211--231, Feb. 2005.

\bibitem{code_Kalantari}
\BIBentryALTinterwordspacing
 [Online]. Available:
  \url{http://www.ece.ucsb.edu/~psen/PaperPages/HDRVideo/HDRVideo_MATLAB_v1.0.zip}
\BIBentrySTDinterwordspacing

\bibitem{code_Mangiat}
\BIBentryALTinterwordspacing
 [Online]. Available:
  \url{http://www.ece.ucsb.edu/~kuochin/release/HDR_MATLAB.zip}
\BIBentrySTDinterwordspacing

\bibitem{reinhard_photographic_2002}
E.~Reinhard, M.~Stark, P.~Shirley, and J.~Ferwerda, ``Photographic tone
  reproduction for digital images,'' \emph{{ACM} Trans. Graphics}, vol.~21,
  no.~3, pp. 267--276, Jul. 2002.

\bibitem{barnes_tog_2009}
C.~Barnes, E.~Shechtman, A.~Finkelstein, and D.~B. Goldman, ``Patchmatch: A
  randomized correspondence algorithm for structural image editing,'' \emph{ACM
  Trans. Graphics}, vol.~28, no.~3, pp. 24:1--11, Jul. 2009.

\bibitem{aydin_tog_2010}
T.~O. Aydin, M.~\v{C}ad\'{\i}k, K.~Myszkowski, and H.-P. Seidel, ``Video
  quality assessment for computer graphics applications,'' \emph{ACM Trans.
  Graphics}, vol.~29, no.~6, pp. 161:1--12, Dec. 2010.

\bibitem{SPIE_Aydin_2008}
T.~O. Ayd{\i}n, R.~Mantiuk, and H.-P. Seidel, ``Extending quality metrics to
  full luminance range images,'' in \emph{Proc. IS\&T/SPIE Human Vision and
  Electronic Imaging}, Feb. 2008, p. 68060B.

\bibitem{TOG_Mantiuk_2011}
R.~Mantiuk, K.~J. Kim, A.~G. Rempel, and W.~Heidrich, ``{HDR-VDP-2}: A
  calibrated visual metric for visibility and quality predictions in all
  luminance conditions,'' \emph{{ACM} Trans. Graphics}, vol.~30, no.~4, pp.
  40:1--40:14, Jul. 2011.

\bibitem{JEI_Narwaria_2015}
M.~Narwaria, R.~K. Mantiuk, M.~P.~D. Silva, and P.~L. Callet, ``{HDR-VDP-2.2}:
  A calibrated method for objective quality prediction of high-dynamic range
  and standard images,'' \emph{J. Electronic Imaging}, vol.~24, no.~1, p.
  010501, Jan. 2015.

\bibitem{SPIC_Narwaria_2015}
M.~Narwaria, M.~P.~D. Silva, and P.~L. Callet, ``{HDR-VQM}: An objective
  quality measure for high dynamic range video,'' \emph{Signal Process.: Image
  Commun.}, vol.~35, pp. 46--60, Jul. 2015.

\bibitem{nie_tip_2002}
Y.~Nie and K.-K. Ma, ``Adaptive rood pattern search for fast block-matching
  motion estimation,'' \emph{IEEE Trans. Image Process.}, vol.~11, no.~12, pp.
  1442--1449, Dec. 2002.

\bibitem{wen_solving_2012}
Z.~Wen, W.~Yin, and Y.~Zhang, ``Solving a low-rank factorization model for
  matrix completion by a nonlinear successive over-relaxation algorithm,''
  \emph{Math. Prog. Comput.}, vol.~4, no.~4, pp. 333--361, Jul. 2012.

\bibitem{hasinoff_noise-optimal_2010}
S.~W. Hasinoff, F.~Durand, and W.~T. Freeman, ``Noise-optimal capture for high
  dynamic range photography,'' in \emph{Proc. {IEEE} Conf. Comput. Vis. Pattern
  Recognit.}, Jun. 2010, pp. 553--560.

\bibitem{granados_optimal_2010}
M.~Granados, B.~Ajdin, M.~Wand, C.~Theobalt, H.-P. Seidel, and H.~P.~A. Lensch,
  ``Optimal {HDR} reconstruction with linear digital cameras,'' in \emph{Proc.
  {IEEE} Conf. Comput. Vis. Pattern Recognit.}, Jun. 2010, pp. 215--222.

\bibitem{magnus1995matrix}
J.~R. Magnus and H.~Neudecker, \emph{Matrix Differential Calculus with
  Applications in Statistics and Econometrics}, 2nd~ed.\hskip 1em plus 0.5em
  minus 0.4em\relax Chichester, UK: John Wiley \& Sons, 1999.

\end{thebibliography}
